\documentclass[preprint,12pt,authoryear]{elsarticle}

\usepackage{amsmath}
\usepackage{float}

\usepackage{graphicx} % more modern
\usepackage{subfigure} 

\usepackage{graphics} %added
\usepackage{siunitx}  %added
\usepackage{lineno}   %added

\usepackage{epsfig}

\usepackage{amssymb}
\journal{Solar Energy}

\begin{document}

%% Add line numbers
%\linenumbers

\begin{frontmatter}



%\title{Benchmarking of four Deep Learning architectures for sky image processing in irradiance forecasting}

\title{Benchmarking of Deep Learning Irradiance Forecasting Models from Sky Images - an in-depth Analysis}
%\title{Benchmarking of Convolutional Architectures for Irradiance Forecasting from Sky Images - An In-Depth Analysis}


%\author[label1]{Quentin Paletta\corref{cor1}}
\author[label1,label3]{Quentin Paletta*}
\author[label3]{Guillaume Arbod}
\author[label1]{Joan Lasenby}

%\author[label1]{Joan Lasenby}

\address[label1]{Department of Engineering, University of Cambridge, UK}
\address[label3]{ENGIE Lab CRIGEN, France}

%\fntext[label1]{Department of Engineering, University of Cambridge, UK}
%\fntext[label3]{ENGIE Lab CRIGEN, France}

%\cortext[cor1]{Corresponding Author at qp208@cam.ac.uk}

%\address{Department of Engineering, University of Cambridge, UK}

\begin{abstract}

{\small A number of industrial applications, such as smart grids, power plant operation, hybrid system management or energy trading, could benefit from improved short-term solar forecasting, addressing the intermittent energy production from solar panels. However, current approaches to modelling the cloud cover dynamics from sky images still lack precision regarding the spatial configuration of clouds, their temporal dynamics and physical interactions with solar radiation. Benefiting from a growing number of large datasets, data driven methods are being developed to address these limitations with promising results. In this study, we compare four commonly used deep learning architectures trained to forecast solar irradiance from sequences of hemispherical sky images and exogenous variables. To assess the relative performance of each model, we used the forecast skill metric based on the smart persistence model, as well as ramp and time distortion metrics. The results show that encoding spatiotemporal aspects of the sequence of sky images greatly improved the predictions with 10 min ahead forecast skill reaching 20.4\% on the test year. However, based on the experimental data, we conclude that, with a common setup, deep learning models tend to behave just as a `very smart persistence model', temporally aligned with the persistence model while mitigating its most penalising errors. Thus, despite being captured by the sky cameras, models often miss fundamental events causing large irradiance changes such as clouds obscuring the sun. We hope that our work will contribute to a shift of this approach to irradiance forecasting, from reactive to anticipatory.}

{\scriptsize *Corresponding author (qp208@cam.ac.uk)}

%\vspace{0.2\baselineskip}
%{\small *Corresponding author (qp208@cam.ac.uk)}

\end{abstract}


%Research highlights


\end{frontmatter}


\section{Introduction}
\label{intro}

Contrary to conventional energy sources, solar panels do not output a stable and controllable energy supply, which is currently limiting their integration into the electric network~\citep{Ela2013}. The short-term variability of solar energy production is mainly caused by the continuous flow of clouds over the facility, sporadically hiding the panels from the sun and reducing the electrical power generated. In addition to statistical time series analysis, a common approach to short-term irradiance forecasting is to understand the spatial and temporal dynamics of the cloud cover from images of the sky \citep{Marquez2013, peng3DCloudDetection2015a}. Besides satellite image analysis~\citep{Perez2013}, the use of ground-taken sky images has gained popularity due to its higher temporal and spatial resolutions, at the cost of a more limited spatial covering~\citep{Chow2011}. From these, the 3D configuration of the cloud cover relative to the ground and the sun can be estimated~\citep{peng3DCloudDetection2014, Blanc2017, kuhnDeterminationOptimalCamera2019}. Given a sequence of images, tracking methods~\citep{Marquez2013, huangCloudMotionEstimation2013, Bernecker2014, Alonso2014, pengHybridApproachEstimate2016, Bone2018} are used to capture the temporal dynamics of the clouds seen by the camera and estimate their current trajectory, which then can be used to predict their future position~\citep{Chow2011} or at least locate incoming clouds in the image~\citep{Quesada-Ruiz2014}. Following this, ray tracing methods can be used to generate an irradiance map on the ground given physical properties of clouds~\citep{nouriDeterminationCloudTransmittance2019} and their position relative to the sun~\citep{Blanc2017, nouriNowcastingDNIMaps2018, nouriCloudHeightTracking2019a}. An in-depth analysis of individual components of the above methods and a review of the current literature on the topic are presented in~\cite{Kuhn2019} and~\cite{Yang2018}, respectively.  However, challenges in irradiance forecasting remain: our perception of the complexity of cloud physical properties from sky images, but also their spatial and temporal dynamics, is still limited~\citep{Brad2002, Nou2018}.

\vspace{1\baselineskip}
Machine learning (ML), increasingly applied across a wide range of fields, might bring novel solutions to hurdles in irradiance forecasting. Current approaches to solar energy prediction focus on a range of supervised and unsupervised learning techniques such as support vector machines, decision trees, \textit{k}-nearest neighbours or gaussian processes~\citep{Voyant2017}. Deep learning (DL), a sub-field of ML, recently gained popularity for its applicability in various areas. Given historical data containing internal and external variables, artificial neural networks (ANNs) can be trained using supervised learning to predict the future irradiance~\citep{Yadav2014}. Used variables range from in situ irradiance measurements, meteorological data, irradiance predictions given by other models or handcrafted features extracted from sky images~\citep{Chu2013a}. ANNs learn patterns from past data, which enables a complex mapping between the input and the output. The training or learning process consists of optimising model parameters to improve predictions on a training set composed of given input-output pairs.

\vspace{1\baselineskip}
More specifically, DL models can be trained for irradiance forecasting by extracting patterns not only from single point data but also from sky images~\citep{pothineni2019b, Feng2020, Wen2020a, palettaConvolutionalNeuralNetworks2020a}. 2D inputs such as photographs can be treated by a specific type of neural network called convolutional neural networks (CNN)~\citep{LeCun1989}). The learning for such models is based on the use of filters trained to find relevant patterns in an image for a given task. The structure of a layered network enables filters from higher layers to use the ability of lower level filters to recognise basic patterns and in turn spot more complex ones. For example, filters from the first layer are able to respond to edges or corners, whereas filters from the last layers might be able to recognise the sun or a specific type of cloud in the sky. In parallel with other past meteorological data, such information can be given to an ANN or another ML model to forecast irradiance. Alternatively, more sophisticated model architectures can be exploited to extract relevant features from a sequence of images: convolutional neural network + long short-term memory network (CNN+LSTM)~\citep{Zhang2018, Siddiqui2019} and 3D convolutional neural network (3D-CNN)~\citep{Zhao2019}.

%\subsection{Contribution}
\vspace{1\baselineskip}
Predicting the interaction of the cloud cover with the sun from a ground camera is difficult, owing to: different types of clouds, lack of visibility of their 3D-conformation from a single distant point of view and complex light-matter interactions. These challenges in irradiance forecasting could be met with the application of DL models, which are unaffected by the difficulties of modelling the cloud cover dynamics with explicit instructions. Moreover, DL approaches have already been shown to outperform hand-crafted methods in many vision-based tasks~\citep{Weinzaepfel2013a, Schroff2015a, Zhang2019, Kwon2019} and, as data-driven methods, they are bound to benefit from the growing number of available datasets and resources~\citep{Pedro2019}.

\vspace{1\baselineskip}
The original contributions of this study are the following:

\begin{itemize}

    \item We conducted a benchmark study on four common deep learning architectures: CNN, CNN+LSTM, 3D-CNN and a convolutional long short-term memory network (ConvLSTM). To a certain extent, these are representative of the current state of the field as they share similar designs: convolutional layers, recurrent layers and densely connected layers trained end-to-end to predict a future irradiance value.

    \item To assess the model's performance, we implemented the \textit{forecast skill} (FS) metric based on the smart persistence model (SPM). This gives a reliable performance score to compare the proposed DL models with other approaches~\citep{Yang2020b}.
    
    \item We evaluate the ability of the different models to predict ramps from a specific metric~\citep{Vallance2017}.
    
    \item We demonstrate, using the \textit{temporal distortion mix}~\citep{Vallance2017}, that the forecasts by the DL models compared in this work, lag in time behind the actual irradiance values. Additionally, we show that these models tend to behave as \textit{very smart persistence} models mitigating, for instance, double errors caused by a forecast temporally misaligned with the actual irradiance.
    
    \item We show that the size of the training set can be more relevant for the ability to generalise irradiance forecasting than the type of DL model.

\end{itemize}

\vspace{1\baselineskip}
The following Sections~\ref{preliminaries} and~\ref{dataset_chapter} outline the irradiance forecasting problem and present the dataset used in this study. The different network architectures benchmarked are presented in Section~\ref{model}. In Section~\ref{results}, forecasting performances of the model are compared to the SPM and assessed through the ramp and distortion metrics. We conclude with a critical discussion of the proposed DL approach.

\section{Preliminaries}
\label{preliminaries}

\subsection{Formulation of the irradiance problem and objectives}
\label{formulation_pb}

Given the diverse applications of solar forecasting, its objectives are varied. However, the main goal of irradiance prediction, given past irradiance measurements ($Y_n$) and external past data ($W_n$), is usually twofold. Firstly, the error between the actual future irradiance $Y_{n+1}$ and the prediction $Y_{n+1}^*$ must be minimised. Secondly, we expect a forecasting model to be able to anticipate sudden irradiance changes caused by the veiling or the unveiling of the sun by a passing cloud (see Section~\ref{ramp_metric})

\vspace{1\baselineskip}

The precision of the forecast can be measured using skill scores based on different metrics such as the mean absolute error (MAE) or the mean squared error (MSE), which quantify the performance of a model relative to a reference (see Section~\ref{forecast_skill}). Historically, the skill score based on the root mean square error (RMSE) is a popular metric in solar forecasting because of its good generalisation properties used to compare methods on different datasets~\citep{Yang2020b}. In practice, the quadratic difference between the prediction $Y_{n}^*$ and the real value $Y_n$ weights larger errors more compared to the MAE.

\subsection{Forecast skill}
\label{forecast_skill}

In addition to the MAE, the MSE and the RMSE, the FS based on the SPM is often used to assess the model's performance. The SPM, which takes into account diurnal changes of the extra-terrestrial irradiance, is specified by Equation~\ref{equ:smart_persistence} below, where $Y_{clr}$ represents the clear sky model estimate of the irradiance and $k_c$ the clear sky index. The database providing the clear sky model estimates used in this study is HelioClim~\citep{Blanc2011}.

\begin{equation}
   Y^*(t+\Delta T) = k_c(t) Y_{clr}(t+\Delta T) \; \; \; \text{with} \; \; \; k_c(t) =  \frac{Y(t)}{Y_{clr}(t)}
   \label{equ:smart_persistence}
\end{equation}

For a given error metric (MAE, MSE), the FS is computed for different forecasting time horizons (e.g. 2 min, 10 min or 20 min) as follows:

\begin{equation}
  FS = \frac{\text{Error}_{SPM} - \text{Error}_{forecast}}{\text{Error}_{SPM}} = 1-\frac{\text{Error}_{forecast}}{\text{Error}_{SPM}}
   \label{equ:FS}
\end{equation}

A negative value of the FS therefore indicates a performance worse than that of the SPM, whereas for positive values, the closer it is to 1, the more accurate the forecast.

\subsection{Ramp metric}
\label{ramp_metric}
Often defined as ramps, sudden irradiance changes caused by clouds sporadically covering the sun, can be predicted from observations of the cloudiness, i.e sky image analysis. The performance of a model on ramp forecasts can be evaluated by the ramp metric presented in~\cite{Vallance2017}. In addition to the mean accuracy of the forecasts, anticipating such events is a key aspect for many industrial applications.

\vspace{1\baselineskip}
The ramp metric is based on the \textit{swinging door} (SD) algorithm used to detect ramps from a time series in wind and solar energy fields. Each ramp segment is defined as a continuous sequence of points following a general trend quantified by its slope. For a given sequence, the ramp ends when one data point falls outside a fixed interval of size $\epsilon$ around the current trend, which delimits an approximation area. More details can be found in the original article by~\cite{Florita2013}.

\begin{figure}%[H]%[ht!]%[h!] 
\centering    
\includegraphics[width=0.7\textwidth]{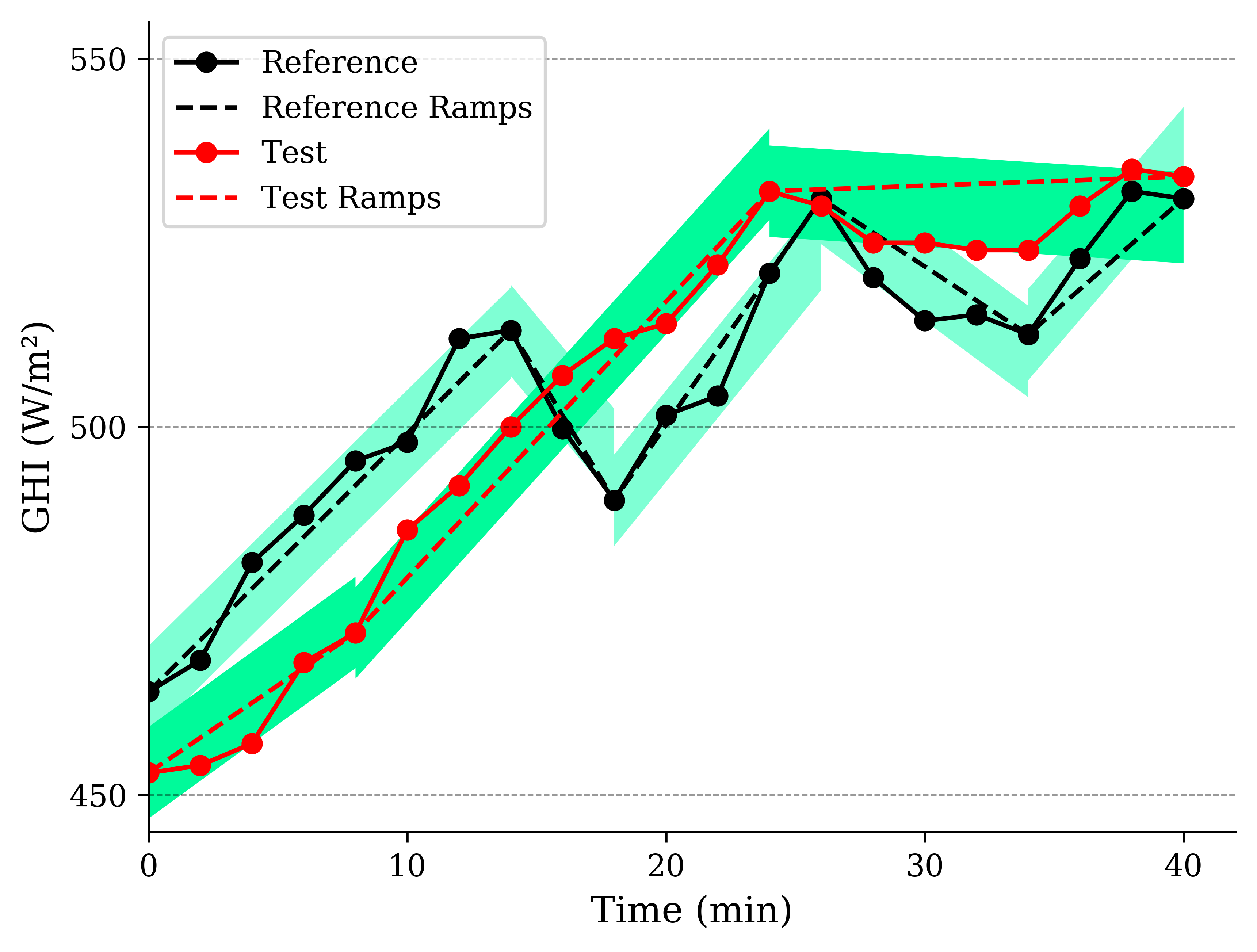}
\caption{Ramps extracted by the SD algorithm from a Test and a Reference series. The shading represents the approximation area for each identified segment.}
\label{fig:ramp_ex_curves}
\end{figure}

\begin{figure}%[H]%[ht!]%[h!] 
\centering    
\includegraphics[width=0.7\textwidth]{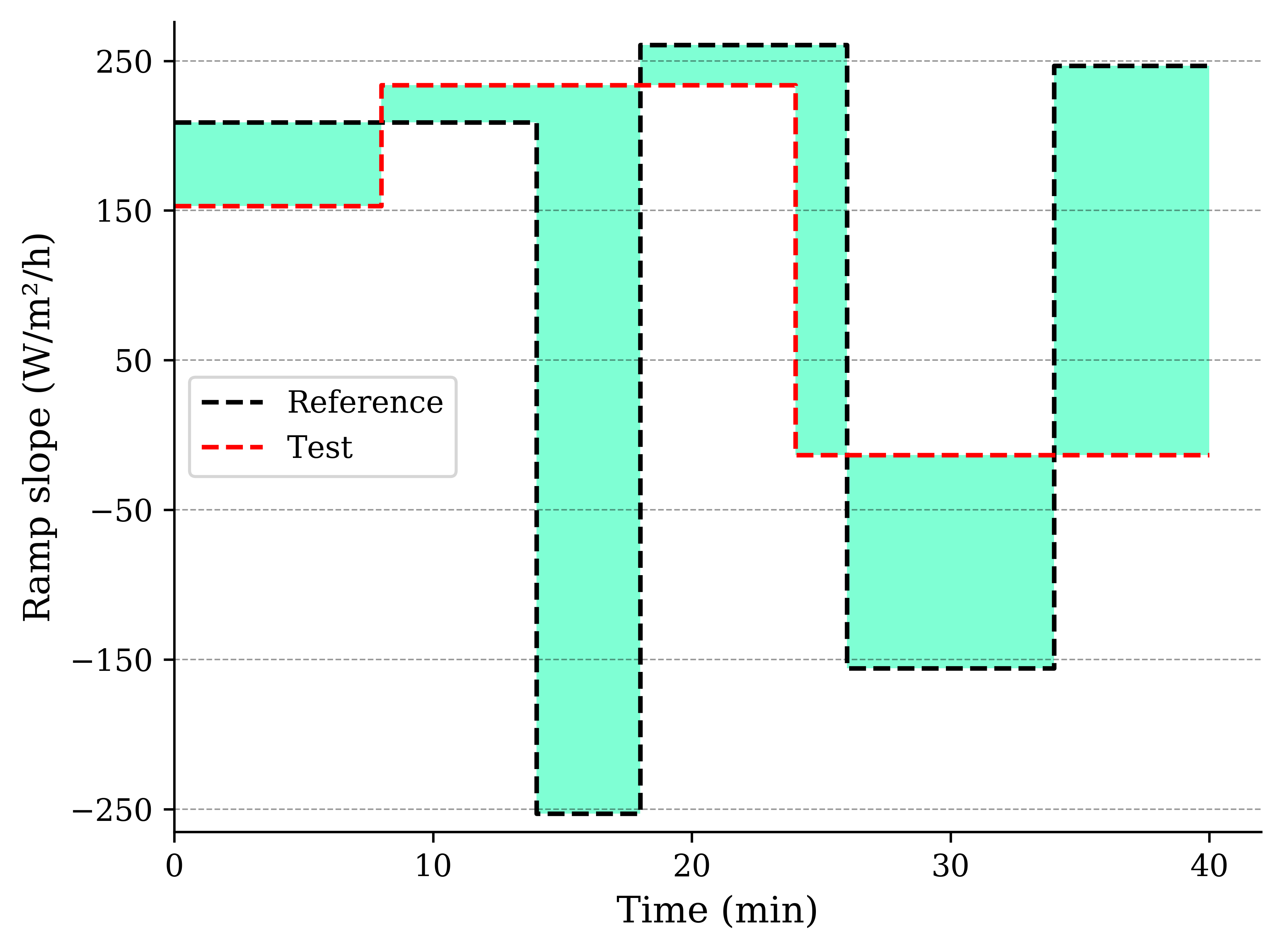}
\caption{Slopes of the ramps detected by the SD algorithm in Figure~\ref{fig:ramp_ex_curves}, inspired by~\cite{Vallance2017}.}
\label{fig:ramp_ex_slopes}
\end{figure}

\vspace{1\baselineskip}
The ramp score defined in Equation~\ref{equ:Ramp}, aims at quantifying the similarity of the ramps extracted by the SD algorithm for two time series R (Reference) and T (Test). This can be inferred by averaging the absolute slope difference at each time step over an interval [$t_{min} : \: t_{max}$]. Figure~\ref{fig:ramp_ex_curves} shows the sequence of ramps extracted from two time series, with their derivatives (SD(T($t$)) and SD(R($t$))) plotted in Figure~\ref{fig:ramp_ex_slopes}. \cite{Vallance2017} suggests defining $\epsilon$ as the product of a fixed parameter $\tau_{CLS}$ (0.05 in this study) and the daily maximum of the irradiance $I_{clr}$ given by a clear sky model to take into account the variability of the solar resource throughout the year (see Equation~\ref{equ:epsilon}).

\begin{equation}
  \text{Ramp Score} = \frac{1}{t_{max}-t_{min}} \int_{t_{min}}^{t_{max}} |\text{SD(T($t$))}-\text{SD(R($t$))}|dt
  \label{equ:Ramp}
\end{equation}

\begin{equation}
  \epsilon = \underset{day \; d}{\tau_{CLS}} \; \text{max} \, I_{clr}
  \label{equ:epsilon}
\end{equation}

\subsection{Temporal distortion index and temporal distortion mix}
\label{tdm}
Assessing the temporal alignment of a forecast with the ground truth is a key aspect of renewable energy forecasting. For instance, predicting an incoming spike with a temporal delay or advancement, could result in a double error impacting the overall performance of the forecasting method.

\begin{figure}%[H]%[ht!]%[h!] 
\centering    
\includegraphics[width=0.65\textwidth]{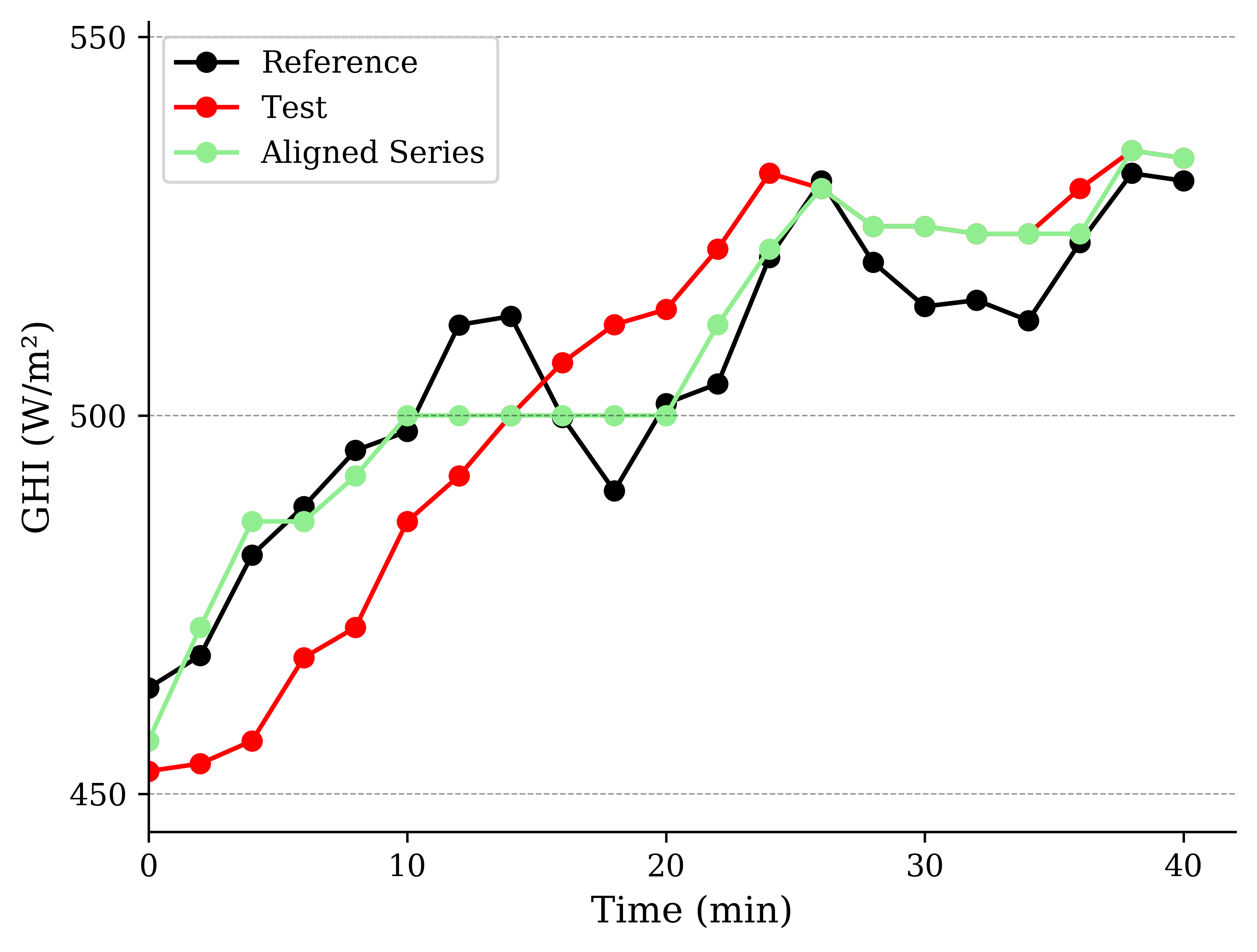}
\caption{Aligned series resulting from the local distortion of the Test series to match the Reference series in both magnitude and time, based on~\cite{Vallance2017}.}
\label{fig:tdi_ex_curves}
\end{figure}

\vspace{1\baselineskip}
The \textit{temporal distortion index} (TDI) was introduced by~\cite{Frias-Paredes2016} to quantify temporal misalignment between two time series. The main principle of this method based on dynamic time warping is to locally find the best distortion to temporally align a Test series with a Reference series (see Figure~\ref{fig:tdi_ex_curves}).

\begin{figure}%[H]%[ht!]%[h!] 
\centering    
\includegraphics[width=0.6\textwidth]{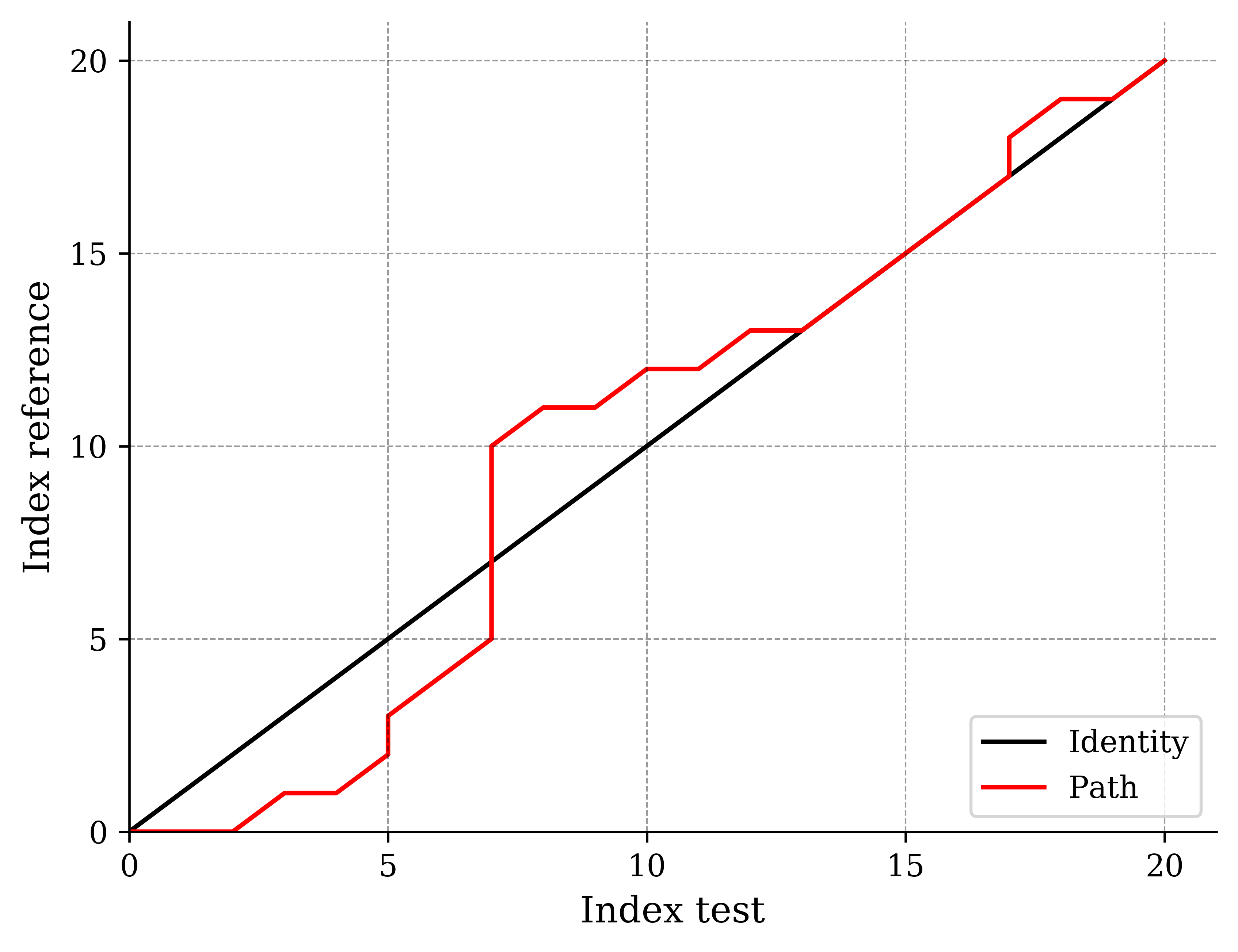}
\caption{Optimal path minimising the time distortion between the Aligned and Reference series, after~\cite{Frias-Paredes2016}.}
\label{fig:tdi_path_ex}
\end{figure}

\vspace{1\baselineskip}
\cite{Frias-Paredes2016} used an optimal path represented in Figure~\ref{fig:tdi_path_ex}, which highlights the misalignment of the two series as indicated by this optimal path deviating from the identity path. The TDI is defined as the area between the optimal and the identity path normalised by the area below the latter, which corresponds to the percentage of temporal distortion relative to the maximal distortion. To disregard the relative scale difference of the two time series considered, we chose to normalise both series independently as a preprocessing step.

\vspace{1\baselineskip}
To convey a more detailed analysis from this method, \cite{Vallance2017} split this index into two components $\text{TDI}_{adv}$ and $\text{TDI}_{late}$ (`in advance' and `late'), corresponding to the area above and below the identity path respectively. Both components can be integrated into a single variable called the temporal distortion mix (TDM), outlining the behaviour of the forecast from -1 (in advance) to 1 (late):

\begin{equation}
  \text{TDM} = 2 \: \frac{\text{TDI}_{late}}{\text{TDI}}-1=1-2 \: \frac{\text{TDI}_{adv}}{\text{TDI}}
  \label{equ:TDM}
\end{equation}

\section{Network training}
\label{dataset_chapter}

\subsection{Dataset}
\label{dataset}
%SIRTA, training and validation set \\

\begin{figure}%[H]%[ht!]%[h!] 
\centering
\begin{minipage}[b]{0.49\textwidth}
    \includegraphics[width=\textwidth]{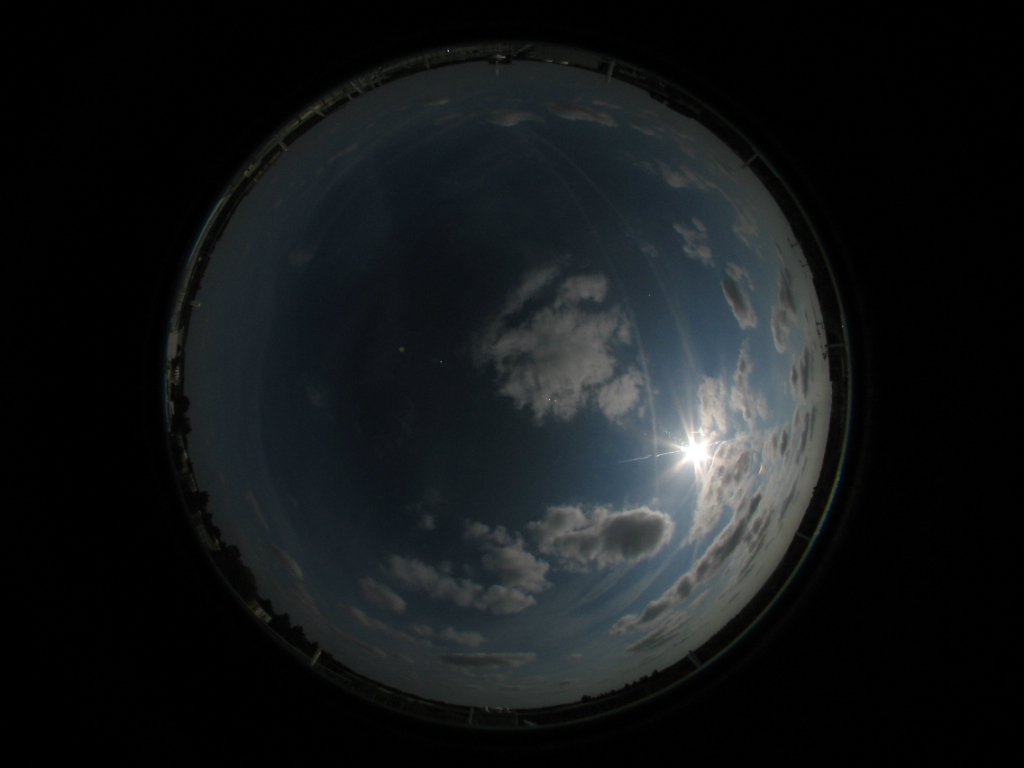}
    %\caption{Short exposure}
    \label{fig:img_sh_2018_7_3_13_0}
  \end{minipage} 
  %\quad
  \begin{minipage}[b]{0.49\textwidth}
    \includegraphics[width=\textwidth]{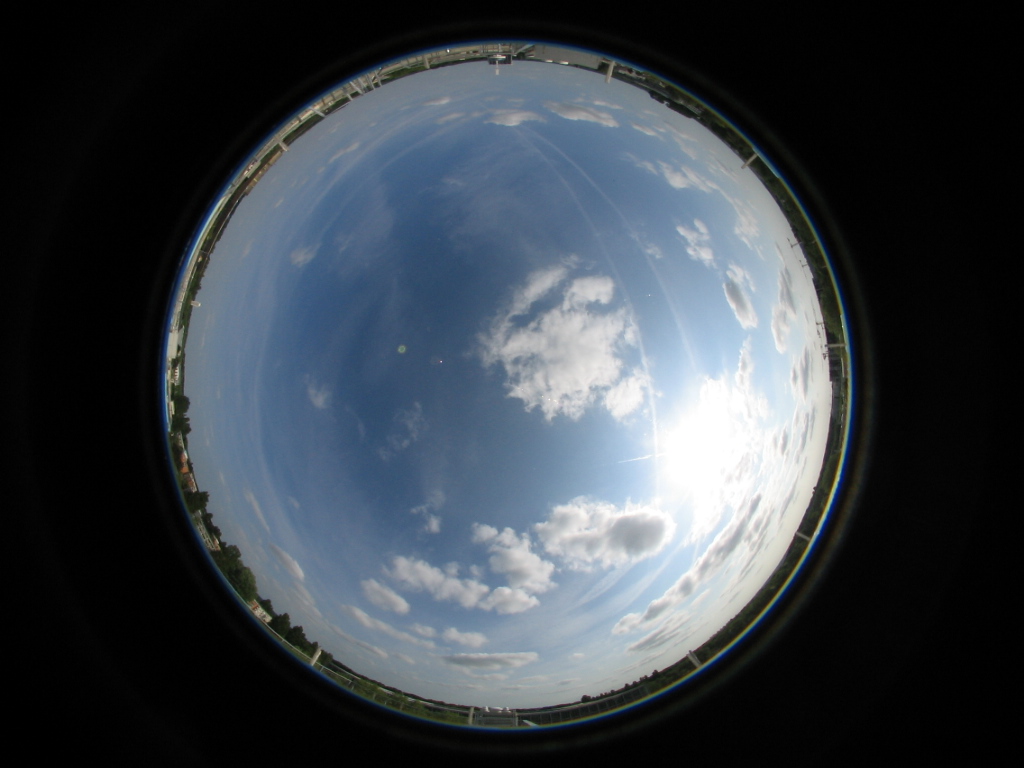}
    %\caption{Long exposure}
    \label{fig:img_sun_204_2018_7_3_13_0}
  \end{minipage}
\caption{Images of the sky taken by a hemispherical camera with short (left panel) and long (right panel) exposures (received from SIRTA laboratory~\citep{sirta}).}
\label{fig:img_1.png}
\end{figure}

The chosen strategy for irradiance forecasting is to train a deep learning (DL) model with sky images taken by a hemispherical camera on the ground (see Figure~\ref{fig:img_1.png}) and a range of auxiliary data such as past irradiance measurements or the angular position of the sun.

\vspace{1\baselineskip}
The dataset used in this study was kindly shared by SIRTA laboratory (Site Instrumental de Recherche par Télédétection Atmosphérique)~\citep{sirta} in France. Samples were collected over a period of three years from 2017 to 2019. The RGB images of the sky were taken by a EKO SRF-02 all-sky camera. Samples are taken every two minutes with a resolution of $768 \times 1024$ pixels and are comprised of two pictures of different exposures referred to as `long' and `short'. The short exposure time (1/2000 sec) provides more details on the region in the proximity of the sun (Left panel on Figure~\ref{fig:img_1.png}), whereas the long exposure (1/100 sec) shifts the focus to the rest of the sky, in particular to distant clouds (Right panel on Figure~\ref{fig:img_1.png}). Additionally, the dataset contains a range of auxiliary data, in particular pyranometer measurements of the global horizontal irradiance (GHI) which are averaged over a minute and the angular position of the sun is available through the solar zenith and azimuthal angles (SZA and SAA).

\vspace{1\baselineskip}
The range of auxiliary data used to train the models at time $t=T$ from samples taken at $t=\{T, T-2, T-4,...\}$ (in minutes) are as follows: GHI($t$), SZA($t$), cos(SZA($t$)), sin(SZA($t$)), SAA($t$), cos(SAA($t$)), sin(SAA($t$)).

\subsection{Training, validation and testing}
\label{training_and_validation}

Samples selected to generate the training and the validation sets were spaced out in time by 4 min and satisfied the constraint that the SZA is lower than 80\si{\degree} based on SIRTA's dataset (more than 10\si{\degree} above the horizon). The distribution of SZA in each set is presented in~\ref{section:dataset_balance}. In addition, the following types of images were filtered out in a quality control preprocessing step: erroneously over-exposed frames or people, birds and insects caught on camera etc. A sample was included in the further analyses if the average pixel intensity of one of its frames, $I_n$, exceeded a threshold value from the previous frame, $I_{n-1}$ (see Equation~\ref{equ:outliers}). The parameter $\epsilon$ was set manually to retain a majority of true positive samples.

\begin{equation}
  \text{Mean}(I_{n}) - \text{Mean}(I_{n-1}) > \gamma \; \text{Mean}(I_{n-1}), \; \; \text{Here $\gamma = 0.1$}
  \label{equ:outliers}
\end{equation}

\vspace{1\baselineskip}
The training set was then generated from 35,000 samples randomly chosen from the 320 available days of 2017 (January to November), the validation set from 10,000 samples from the 320 available days of 2018 (Mid-February to December with 9 consecutive missing days in September) and the test set from 10,000 samples from the 363 available days of 2019 (January to November). The distribution of months in each set is presented in~\ref{section:dataset_balance}.

\vspace{1\baselineskip}

A similar procedure is applied to generate the TDM training, validation and test sets. A sequence sample is made of 100 consecutive samples corresponding to 3 h 18 min, the last sample meeting the previous condition on the SZA. Each set is made of a hundred of such sequences randomly chosen among all possible sequences available from its corresponding year with 30 min between each sequence.

\vspace{1\baselineskip}
The loss functions used by the model as a reference to assess its own performance are the regularised MSE and MAE defined in Equations~\ref{equ:MAE} and~\ref{equ:MSE}, with the weight decay $\lambda$ and the set of regularisation parameters \{$w_j$\}.

\begin{equation}
  L_1(\{Y^*\}, \{Y\}) = \mbox{MAE}(\{Y^*\}, \{Y\}) = \frac{1}{n} \sum_{i=1}^{n} |\mbox{Y}^*_{k} - \mbox{Y}_{k}| + \frac{1}{2} \lambda \sum_{j=1}^{d} |w_j|^2
   \label{equ:MAE}
\end{equation}

\begin{equation}
  L_2(\{Y^*\}, \{Y\}) = \mbox{MSE}(\{Y^*\}, \{Y\}) = \frac{1}{n} \sum_{i=1}^{n} (\mbox{Y}^*_{k} - \mbox{Y}_{k})^2 + \frac{1}{2} \lambda \sum_{j=1}^{d} |w_j|^2
   \label{equ:MSE}
\end{equation}

\section{Model architectures}
\label{model}

\subsection{General approach}
\label{general_approach}

\begin{figure}%[H]%[ht!]%[h!] 
\centering    
\includegraphics[width=1.1\textwidth]{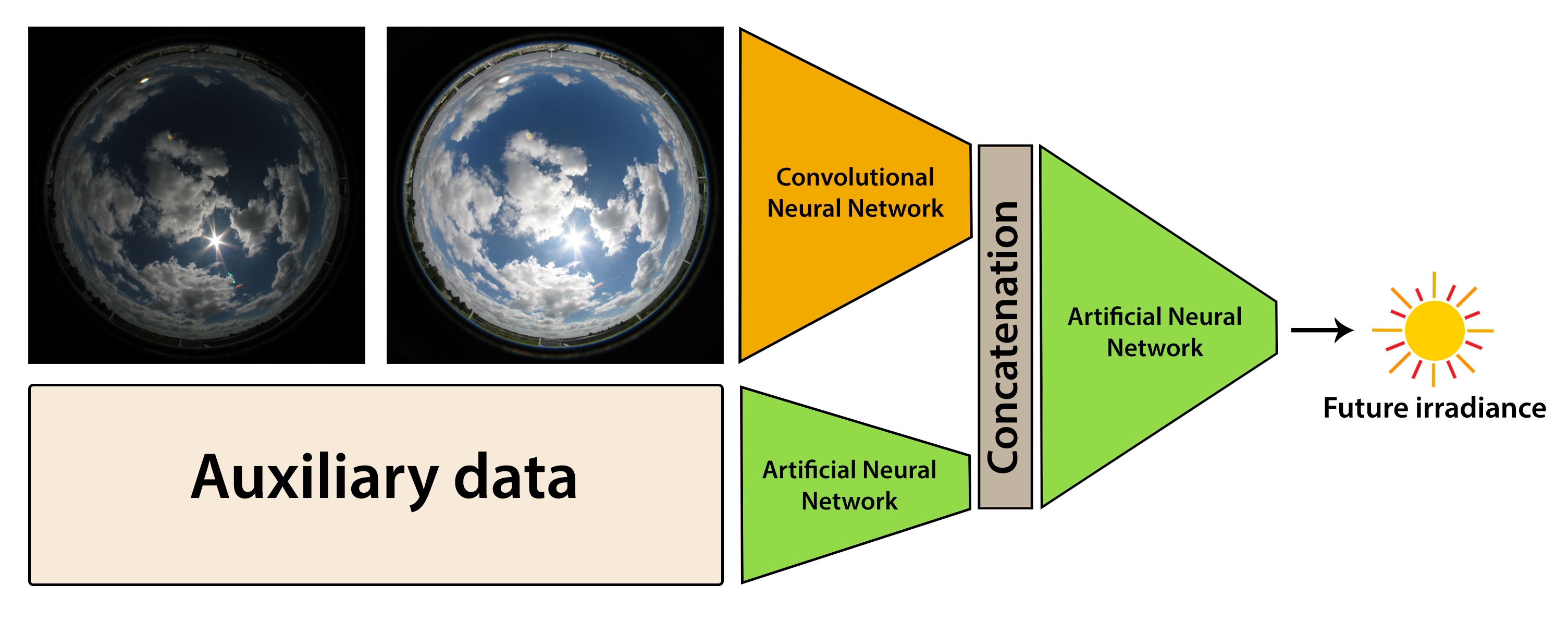}
\caption{General architecture used to forecast irradiance from available sky images and auxiliary data. Two parallel neural networks are trained to encode both types of data into a common latent space. Extracted features are then fed into an ANN, which outputs the future irradiance estimate.}
\label{general_model}
\end{figure}

The four DL approaches we propose and compare in this study are presented in the following subsections. The general method depicted in Figure~\ref{general_model} consists of two parallel networks used to encode the auxiliary data and the images respectively, the image encoder being the main focus of this study. Both networks are merged into a single artificial neural network which outputs the irradiance forecast. The resulting model is trained end-to-end through back propagation. The approach is not iterative, hence the model is trained independently for each forecast horizon. Detailed architectures are listed in~\ref{section:architectures}.

\vspace{1\baselineskip}
Applied data preprocessing was consistent between all four models. Auxiliary data were normalised to retain their respective scale homogeneity. Images were cropped and downscaled through bilinear filtering from $1024 \times 768$ pixels to a resolution of $128 \times 128$ pixels based on the best forecast skill obtained for the CNN model, confirming an experiment by~\cite{Sun2018a}: the higher the resolution, the higher the performances until it reaches a plateau. Transforming each RGB image into a single grey scale image was done for computing performance reasons. Training such models on RGB images may lead to higher performances at the cost of a longer training time.

\subsection{Hyperparameter tuning}
%\label{section:hyperparameters}

The design of each architecture is based on the ResNet architecture~\citep{He2016} used in many computer vision applications. Model parameters are optimized by the adaptive moment estimation ADAM~\citep{Kingma2015}. Given the training time of several hours for each model on a GPU, hyperparameters were mostly manually tuned through grid search based on the forecast skill value for the 10 min ahead forecast. The main parameters tuned are as follows: number of ResNet units, number of convolutional layers with a stride of 2, number of filters or nodes, dropout rate for the final dense layers, learning rate, weight decay and batch size. The activation function for both densely connected layers and convolutional layers is the rectified linear unit (ReLU)~\citep{Nair2010}. Also, the context given to all four models was set the same to 8 minutes. Increasing the size of the sequence further proved to be costly in terms of training time with a limited performance increase. The outcome of the tuning is depicted in Table~\ref{table:Hyperparameters_table} and in~\ref{section:architectures}. Another promising approach is to evaluate the model performance on randomly chosen sets of parameters. The reader may refer to~\cite{Venugopal2019} for a possible implementation.

\renewcommand{\arraystretch}{1.1}
\begin{table}%[H]
%\vskip 0.15in
\caption{Hyperparameters and features of the models.}
\begin{center}
\begin{small}
%\begin{sc}
\begin{tabular}{lccccc}
%\hline
 & Range & CNN & CNN+LSTM & 3D-CNN & ConvLSTM\\
%\cline{2-4}
%& MSE & RMSE & MAE \\
\hline
Learning rate & [$10^{-2}$, $10^{-7}$] & $10^{-4}$ & $10^{-4}$ & $10^{-5}$ & $10^{-5}$ \\
Weight decay & [$0$, $10^{-7}$] & $10^{-5}$ & 0 & $10^{-6}$ & $10^{-6}$\\
Dropout & [0, 0.9] & 0 & 0 & 0.1 & 0 \\
%2D Dropout & 0 & 0 & 0.25 & \\
Batch size & [5, 256] & 10 & 10 & 10 & 10 \\
\\
\hline
Indicative Training time & & 5h30 & 7h30 & 8h & 7h \\
(NVIDIA TITAN Xp)\\
Number of param. & & 0.4 M & 12.6 M & 9.4 M & 4.4 M \\
\hline
\end{tabular}
%\end{sc}
\end{small}
\end{center}
\vskip -0.1in
\label{table:Hyperparameters_table}
\end{table}

\subsection{Convolutional neural network model (CNN)}
\label{cnn}

\begin{figure}[ht]%[H]%[ht!]%[h!] 
\centering    
\includegraphics[width=1.2\textwidth]{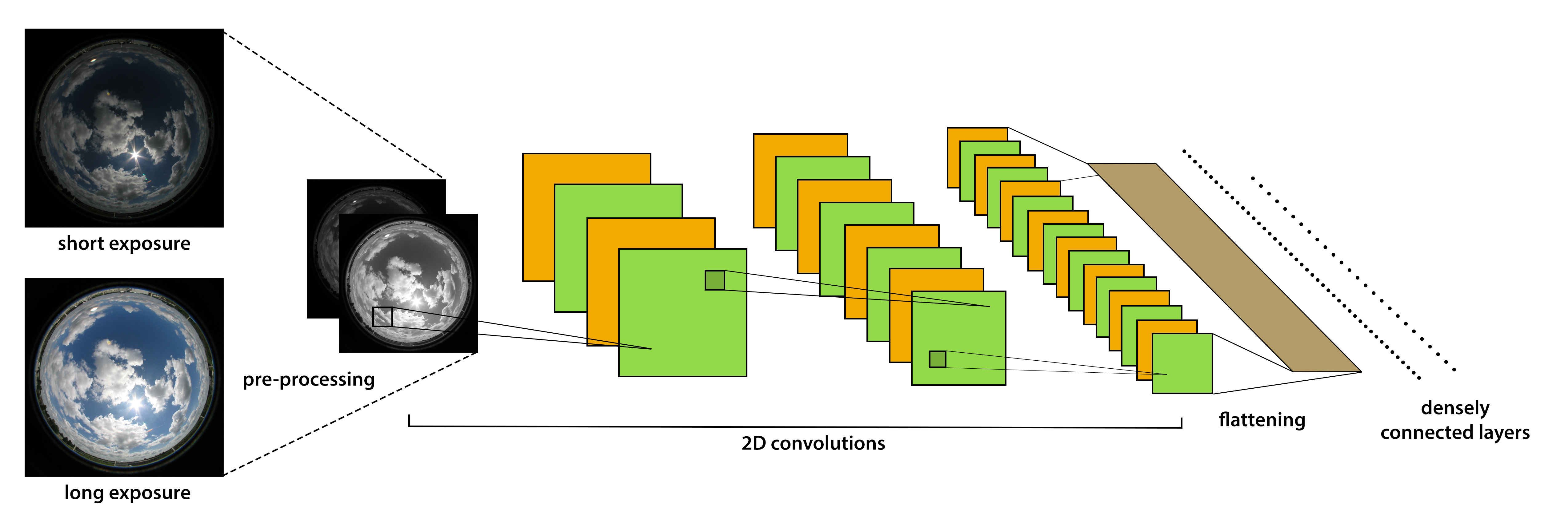}
\caption{CNN image encoder.}
\label{fig:encoder_CNN}
\end{figure}

The `CNN model' consists of two distinct networks merged into a single network made of densely connected layers. The image encoder for this architecture is outlined in Figure~\ref{fig:encoder_CNN}. Post-processed images are convolved with an increasing number of filters through the network. Convolutions with a stride of 2 are used to further downscale the dimension of the data. After flattening the 2D output of the set of convolutional layers, a latent vector of lower dimension is obtained through a set of densely connected layers of a decreasing size. A more detailed model architecture is presented in ~\ref{section:architectures}.

\vspace{1\baselineskip}
Although not designed to encode sequences of images, a limited context of past images can be provided to the image encoder. In this study, the CNN model was given five pairs of images, long and short exposures, taken every two minutes from time $t$ to $t-8\;\text{min}$.

\subsection{Convolutional neural network + long short-term memory model (CNN+LSTM)}
\label{cnn+lstm}

One drawback of the previous architecture is its limited ability to encode temporal information from time series data. However, this can be achieved through a range of recurrent neural networks such as the long short-term memory (LSTM) network~\citep{Hochreiter1997}; this particular design proved to be efficient in avoiding vanishing gradients.

\vspace{1\baselineskip}
An LSTM network uses a hidden state (cell) to keep track of the past information. The cell consists of a 1D representation of the past sequence updated after each time step. The output of the LSTM network combines information from both the current sample and the cell state. It is trained end-to-end by taking a sequence of 1D vectors as an input and outputting a 1D representation of the given sequence.

\begin{figure}[ht]%[H]%[ht!]%[h!] 
\centering    
\includegraphics[width=1.2\textwidth]{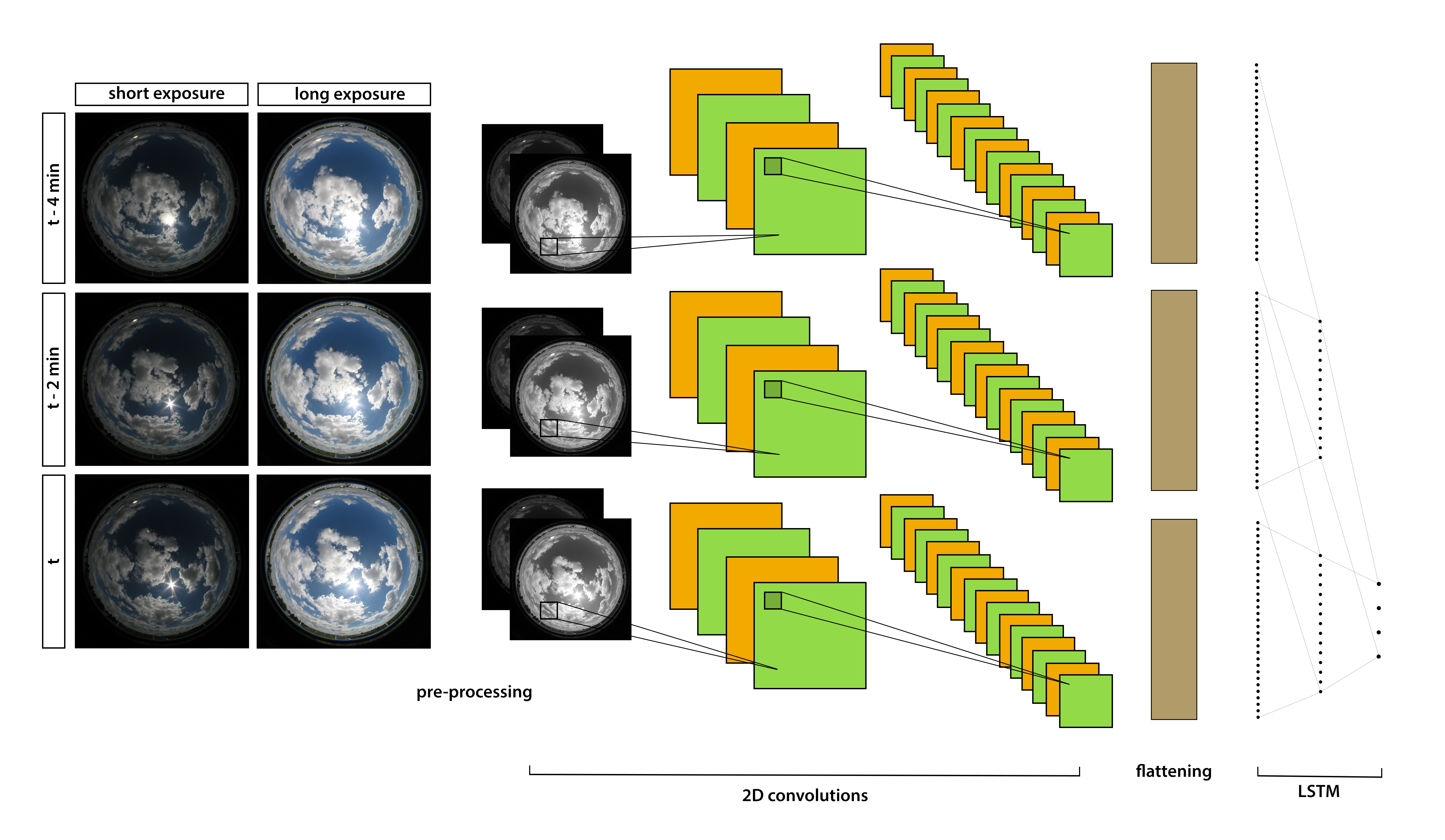}
\caption{LSTM image sequence encoder.}
\label{fig:encoder_LSTM}
\end{figure}

\vspace{1\baselineskip}
To encode a sequence of 2D input such as images through a LSTM network, images have to be first encoded into a 1D representation. As depicted in Figure~\ref{fig:encoder_LSTM}, the pairs of images are first processed in parallel by a single CNN architecture (shared weights) before their one dimensional representation is sequentially sent to the LSTM network.

\subsection{3D convolutional neural network (3D-CNN)}
\label{3dconv}

\begin{figure}[ht]%[H]%[ht!]%[h!] 
\centering    
\includegraphics[width=1.2\textwidth]{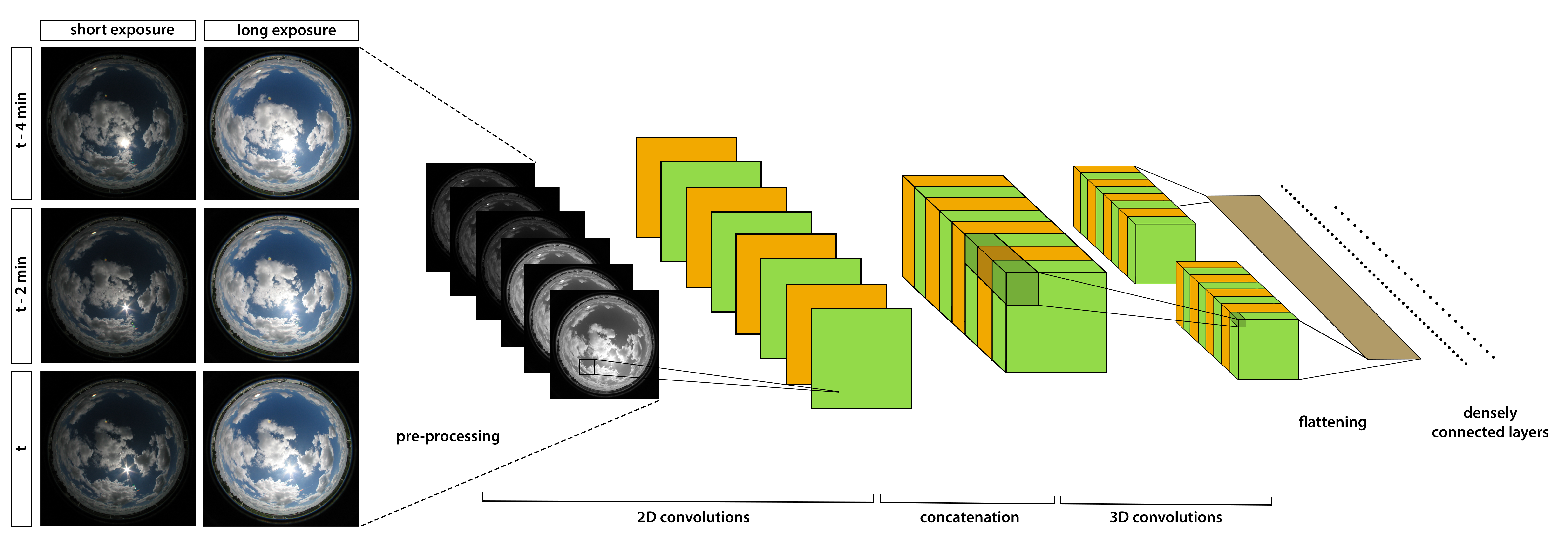}
\caption{3D-CNN image sequence encoder.}
\label{fig:encoder_3Dconv}
\end{figure}

In addition to the two spatial dimensions of an image, 3D-CNN networks integrate the temporal dimension. The input is a sequence of images concatenated after preprocessing into a 3D block where a convolution is applied (Figure~\ref{fig:encoder_3Dconv}).

2D convolutions with a stride of 2 are first applied along the X and Y axes to sub-sample the images from $128 \times 128$ to $8 \times 8$ pixels for instance, before further spatiotemporal dimension reduction. Similarly to a 2D image sequence encoder, the output of the convolutional layers is fed flattened into a vector fed into a set of densely connected layers.

\vspace{1\baselineskip}
In parallel, the sequence of corresponding auxiliary data is encoded using 1D LSTM units, akin to the LSTM model but prior to the merging with the image sequence encoder (see~\ref{section:architectures}).

\subsection{Convolutional long short-term memory model (ConvLSTM)}
\label{convlstm}

The convolutional long short-term memory model was developed by~\cite{Shi2015} to adapt the 1D LSTM architecture for 2D inputs. It was initially applied successfully for short-term precipitation forecasting from satellite images and was able to capture spatiotemporal patterns better than the traditional LSTM.

\begin{figure}[ht] %[H]%[ht!]%[h!] 
\centering    
\includegraphics[width=1.2\textwidth]{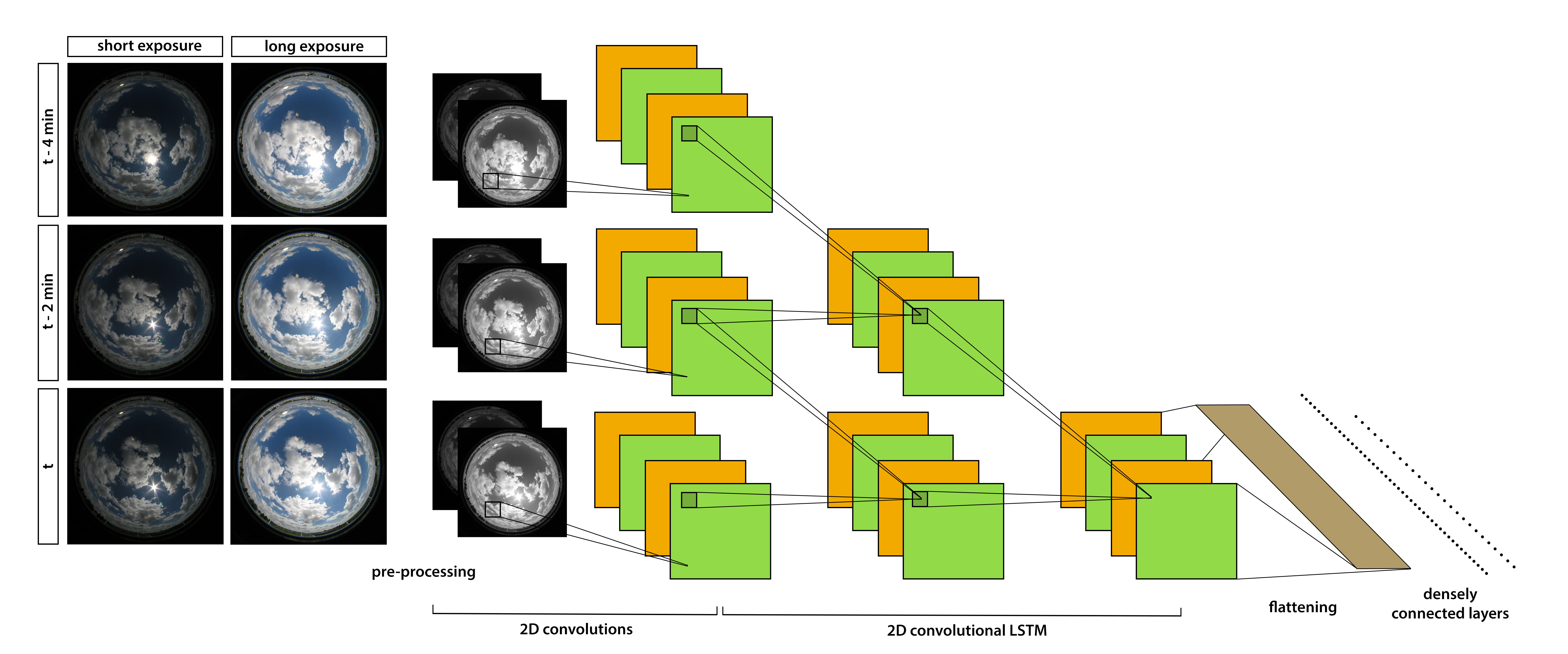}
\caption{ConvLSTM image sequence encoder.}
\label{fig:encoder_Convlstm}
\end{figure}

Similarly to the 3D-CNN model and as suggested by~\cite{Shi2015}, spatial features are first extracted through 2D convolutions and the corresponding feature maps given to the convolutional LSTM units (see Figure~\ref{fig:encoder_Convlstm}). The output is then flattened and fed into dense layers. The auxiliary data encoder is the same as in the 3D-CNN architecture, i.e. dense layers are followed by a LSTM unit (see~\ref{section:architectures}).

\section{Results}
\label{results}

\subsection{Forecasting performance}
\label{overall_performance}

The performances on 10 min ahead forecasting of the four different architectures optimised for both the $L_1$ and the $L_2$ loss functions are presented in Table~\ref{results_table}.  One can see that the ConvLSTM  and the 3D-CNN model achieve the best performances on the forecast skill metrics with 20.4\% and 19.7\% improvement relative to the SPM, respectively. Overall, models including recurrent neural networks (LSTM, 3D-CNN and ConvLSTM) perform better than the one that does not encode temporal patterns (CNN). In particular, they perform significantly better on the ramp score with more than 25\% improvement.

\renewcommand{\arraystretch}{1.1}
\begin{table}[H]
\caption{Metric scores of the models included in the benchmark study for different loss functions (10 min ahead forecast).}
%\vskip 0.15in
\begin{center}
\begin{small}
%\begin{sc}
\begin{tabular}{lccccccccc}
%\hline
 & & \multicolumn{3}{c}{Forecast Skills [\%] $\nearrow$} & Ramp & Quantile (95\%) & TDI & TDM \\
\cline{3-5}
Models & Loss & MSE & RMSE & MAE & [W/$\text{\lowercase{m}}^2$/min] & [W/$\text{m}^2$] & [\%] & $\in[-1, 1]$ \\
\hline
Smart Pers. & - & 0 & 0 & 0 & 28.9 & 349.7 & \textbf{7.8} & 0.79 \\
\\
CNN & $L_1$ & 32.9 & 18.1 & 6.7 & 20.7 (\textit{-28.4\%}) & 279.9 (\textit{-20.0\%}) & 9.4 & 0.48 \\
 & $L_2$ & 32.9 & 18.1 & -3.7 & 20.2 (\textit{-30.1\%}) & \textbf{273.6 (\textit{-21.8\%})} & 10.5 & 0.34 \\
\\

LSTM & $L_1$ & 32.1 & 17.6 & 10.0 & 21.1 (\textit{-27.0\%}) & 280.4 (\textit{-19.8\%}) & 9.4 & 0.49 \\
 & $L_2$ & 34.8 & 19.2 & 3.1 & 20.2 (\textit{-30.1\%}) & 275 (\textit{-21.4\%}) & 10.0 & \textbf{0.34} \\
\\
3D-CNN & $L_1$ & 32.6 & 17.9 & \textbf{10.4} & 21.6 (\textit{-25.3\%}) & 280.9 (\textit{-19.7\%}) & 9.1 & 0.61 \\
 & $L_2$ & 35.5 & 19.7 & 5.8 & \textbf{19.6 (\textit{-32.2\%})} & 274.3 (\textit{-21.6\%}) & 9.4 & 0.49 \\
\\
ConvLSTM & $L_1$ & 33.8 & 18.7 & 9.4 & 20.9 (\textit{-27.7\%}) &  278.2 (\textit{-20.4\%}) & 8.9 & 0.59 \\
 & $L_2$ & \textbf{36.6} & \textbf{20.4} & 6.2 & 20.2 (\textit{-30.1\%}) & 274.1 (\textit{-21.6\%}) & 9.8 & 0.64 \\
\hline
\end{tabular}
%\end{sc}
\end{small}
\end{center}
%\vskip -0.1in
\label{results_table}
\end{table}

\begin{figure}[H]%[ht!]%[h!] 
\centering    
\includegraphics[width=1.1\textwidth]{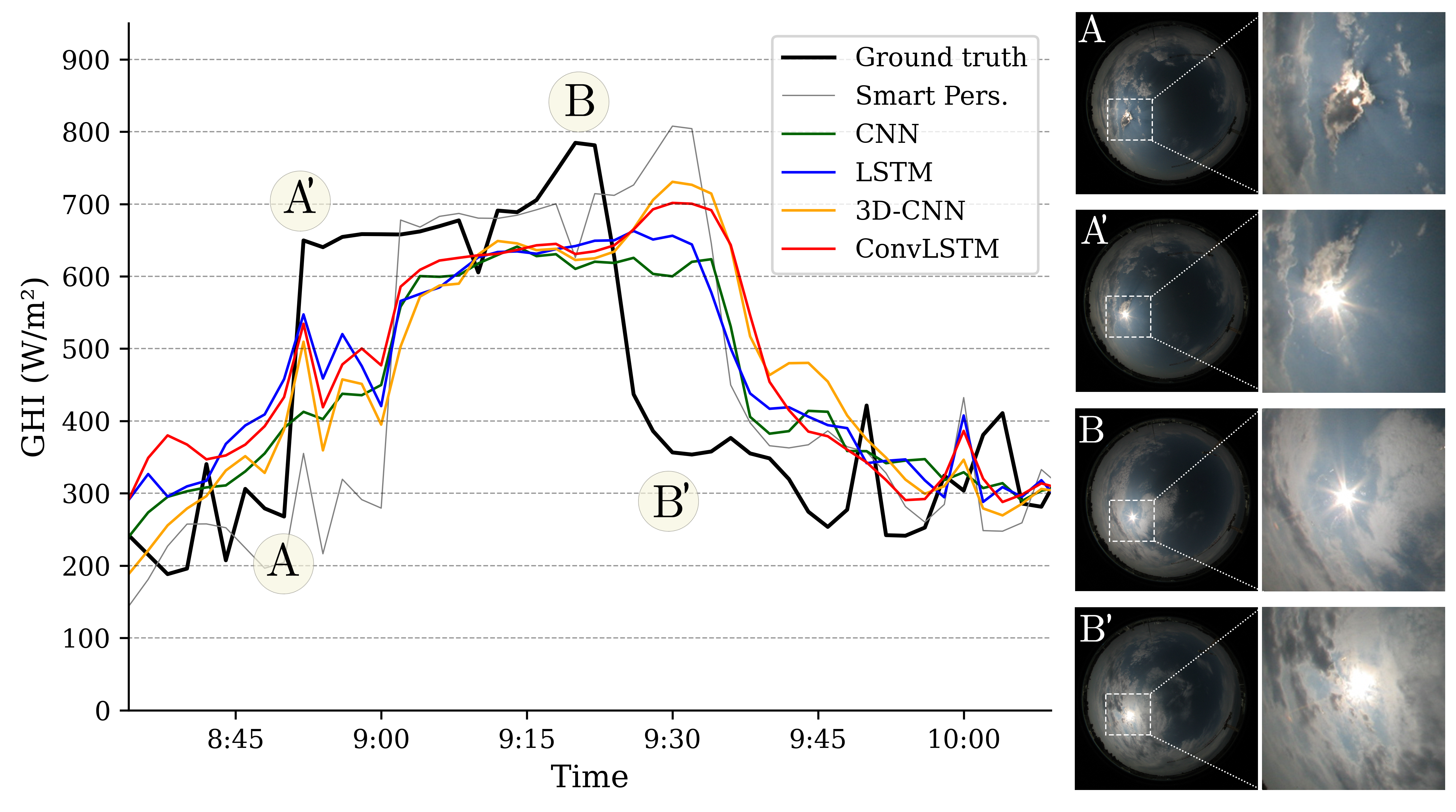}
\caption{Prediction curve comparison for the 10 min ahead forecast ($\text{L}_2$ loss) (plotted for data from 26/07/2019). Despite marginal differences, all four models seem to follow the same behaviour.}
\label{fig:time_series_benchmark_4}
\end{figure}

Regarding time distortion between prediction series and the ground truth, the TDI does not draw a clear difference between models with an index ranging from 8.7\% to 10\%, noticeably higher than the SPM (7.8\%). However, the $L_2$ loss function leads to a higher distortion compared to the $L_1$ loss for three out of the four models. Also, it is worth noticing that the predictions of the DL models are regularly behind the ground truth with a TDM over 0.30 as highlighted in Figure~\ref{fig:time_series_benchmark_4}. All models predict a peak between 9:30 and 9:35 am, temporally aligned with the persistence model prediction, i.e. 10 min after the actual event. This indicates that, with the proposed settings, DL models frequently miss critical events. In many cases, they all act similarly as a `very smart persistence' model increasing their respective skill score by marginally improving over the SPM based on past data at the cost of time delay.

\begin{figure}[H]%[ht!]%[h!] 
\centering    
\includegraphics[width=1.1\textwidth]{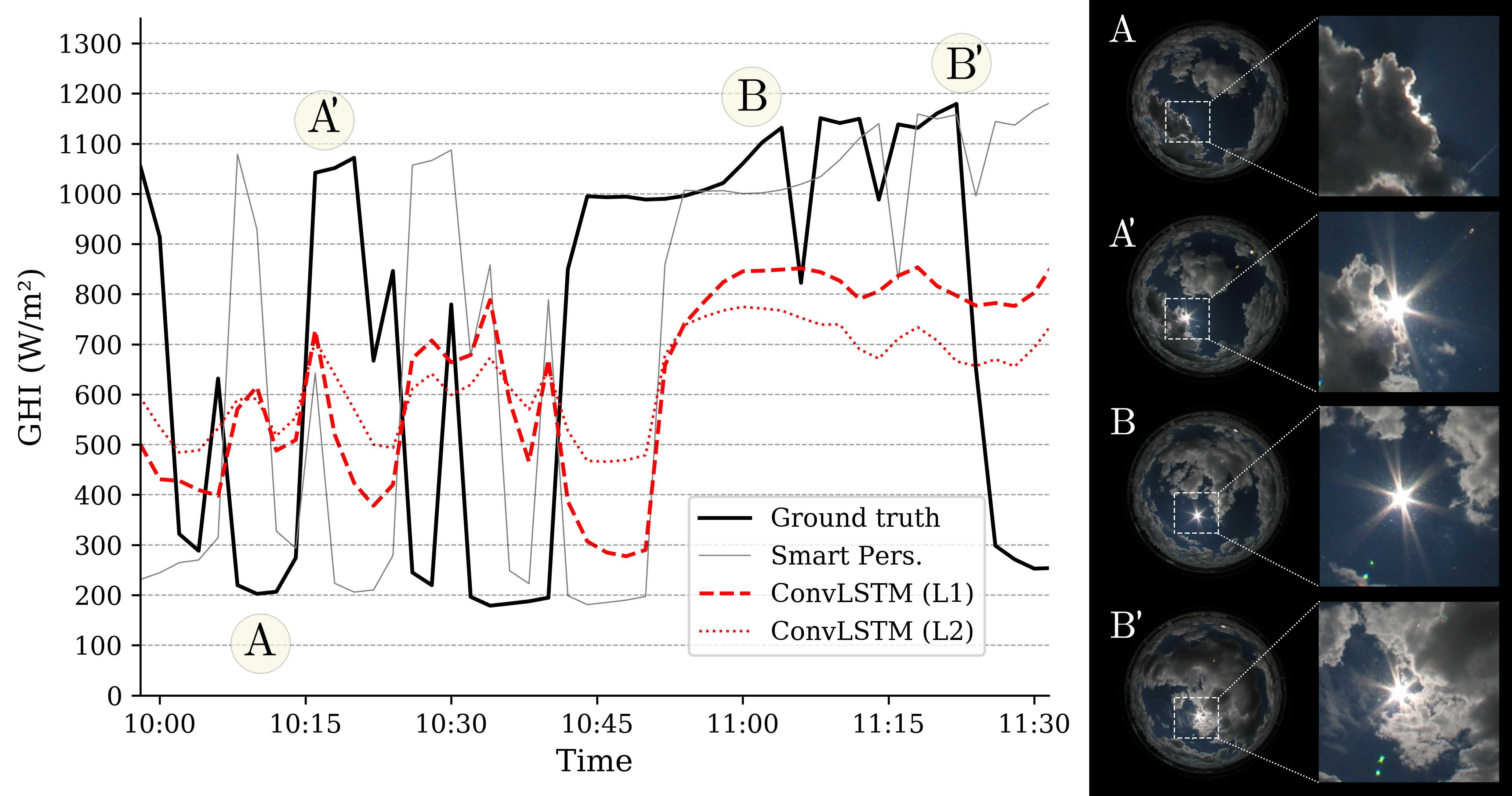}
\caption{10 min ahead prediction curves comparison for different loss functions (plotted for data from 29/05/2019). Penalising larger errors more with the $\text{L}_2$ loss function lowers the variability of the forecast.}
\label{fig:time_series_benchmark_1}
\end{figure}

\begin{figure}[ht]%[H]%[ht!]%[h!] 
\centering
\begin{minipage}[b]{0.49\textwidth}
    \includegraphics[width=\textwidth]{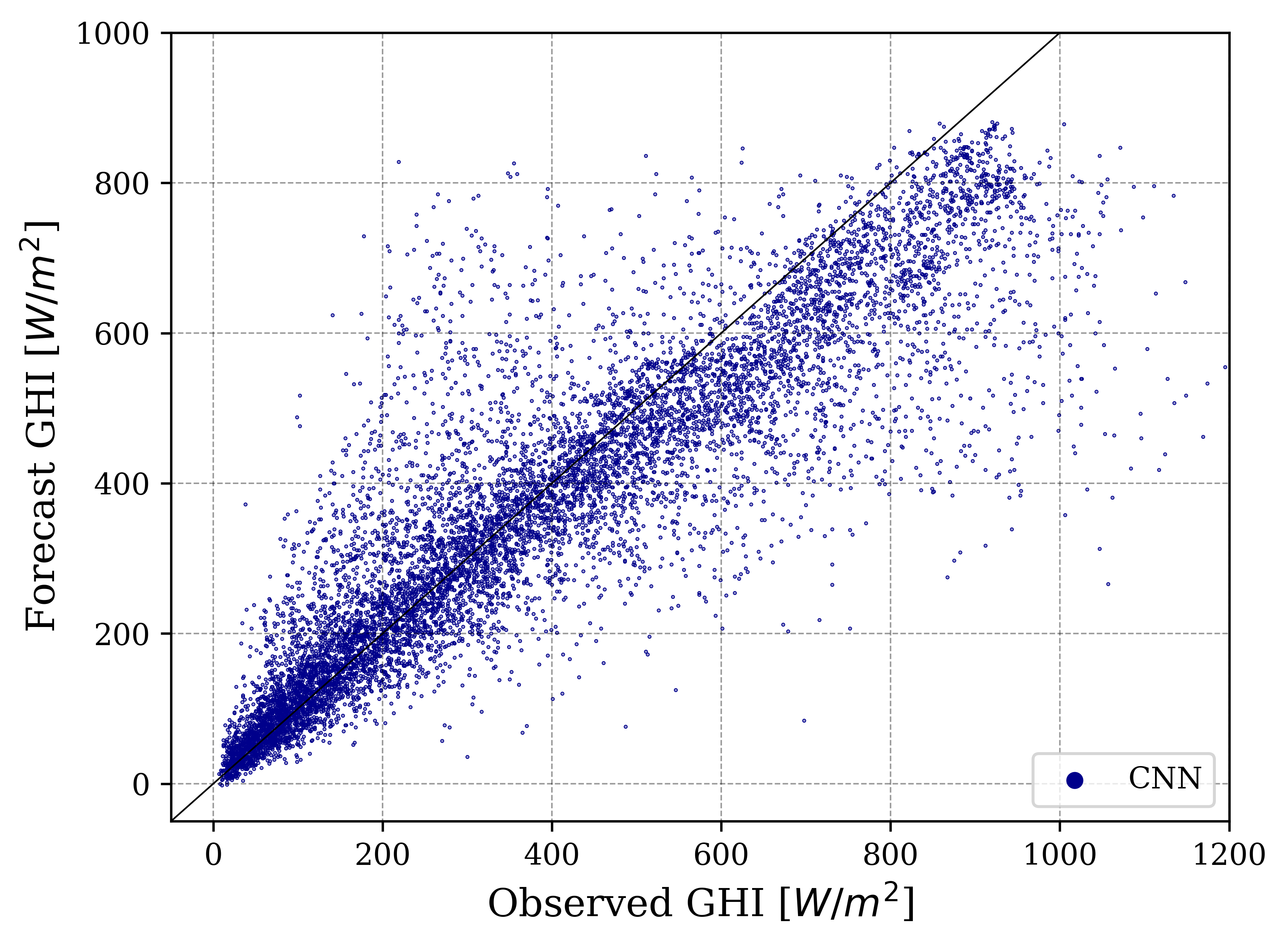}
    %\caption{Short exposure}
    \label{fig:scatter_plot_cnn}
  \end{minipage} 
  %\quad
  \begin{minipage}[b]{0.49\textwidth}
    \includegraphics[width=\textwidth]{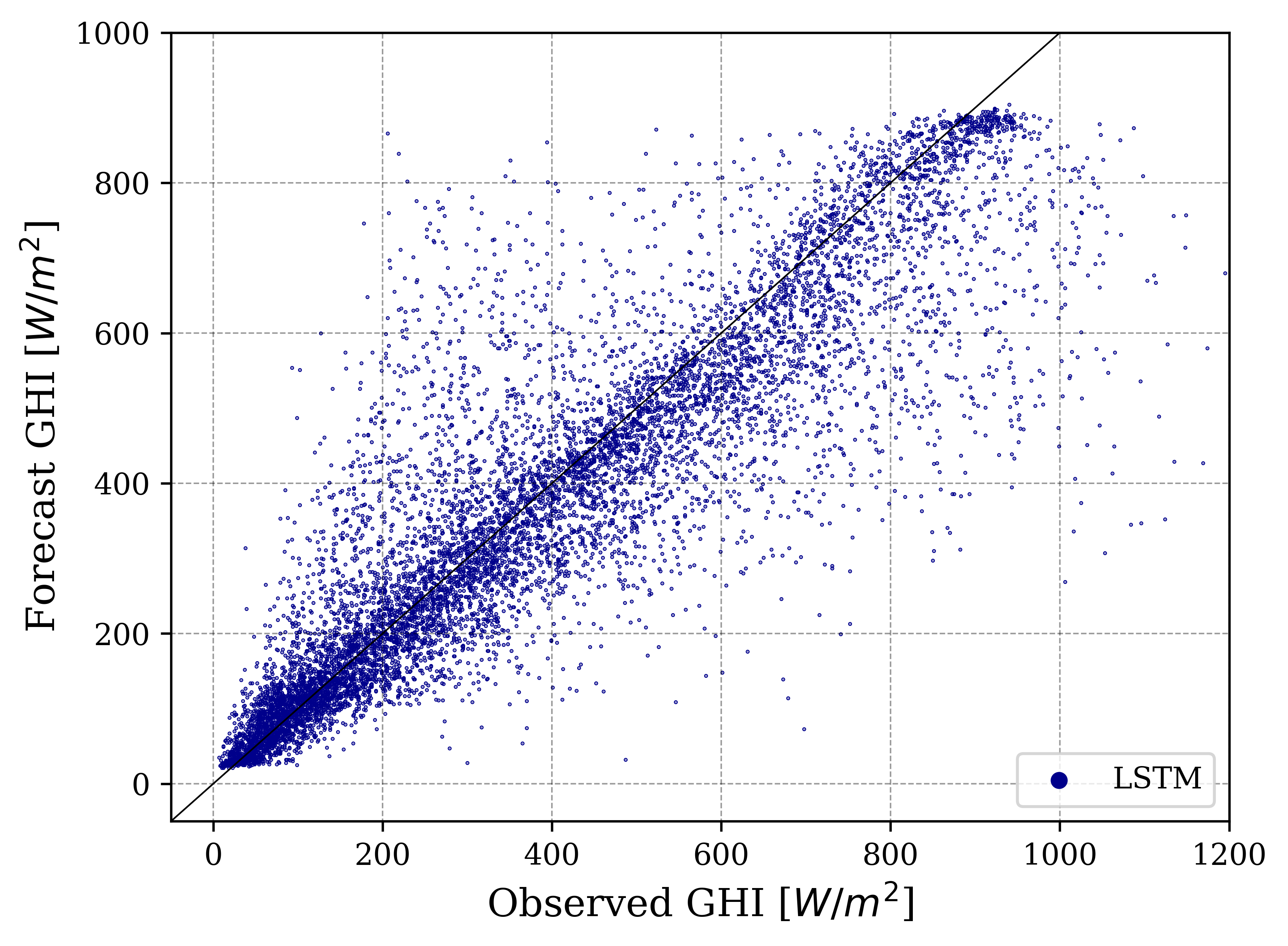}
    %\caption{Long exposure}
    \label{fig:scatter_plot_lstm}
  \end{minipage}
  
 \begin{minipage}[b]{0.49\textwidth}
    \includegraphics[width=\textwidth]{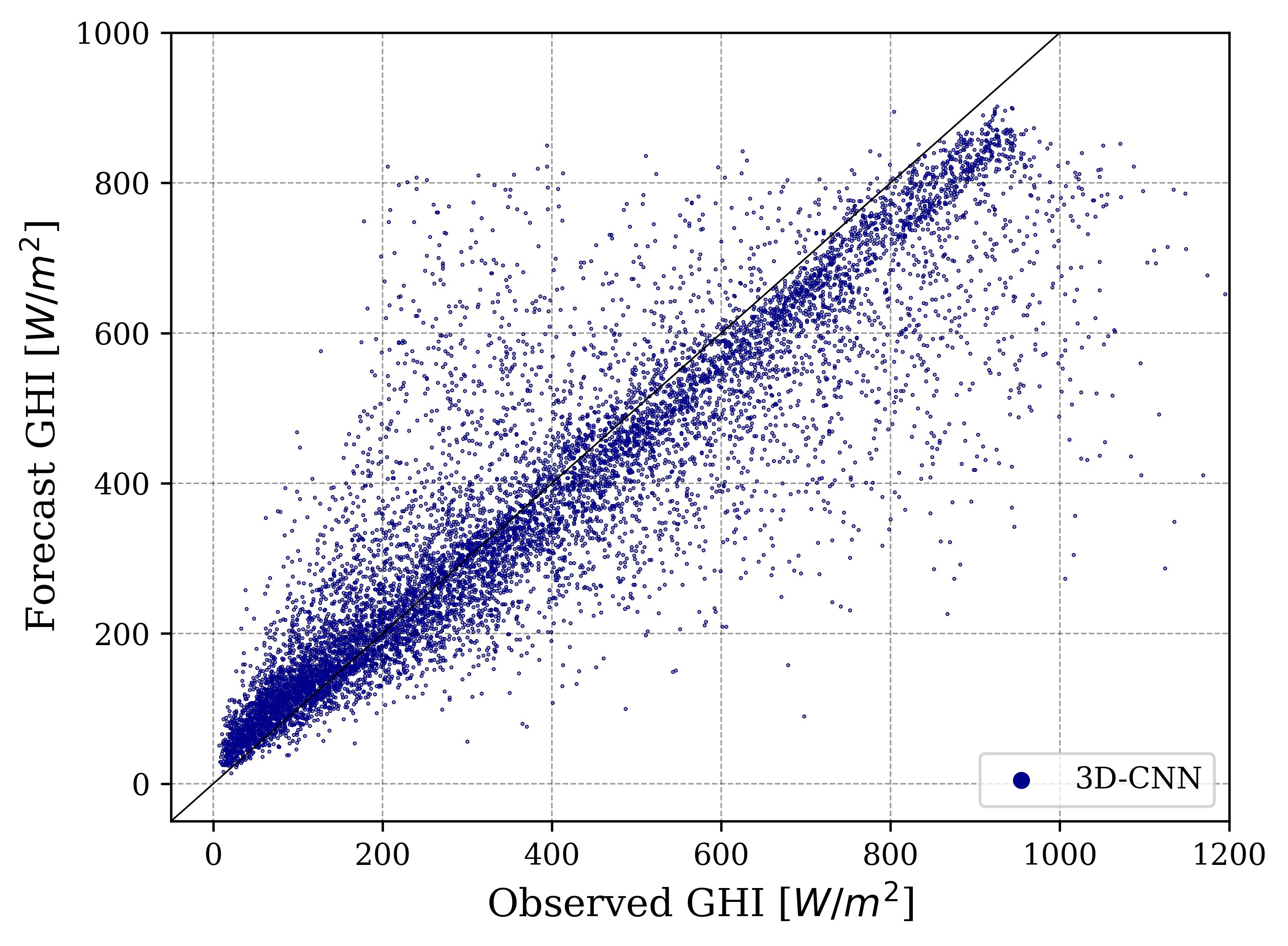}
    %\caption{Short exposure}
    \label{fig:scatter_plot_3dconv}
  \end{minipage} 
  %\quad
  \begin{minipage}[b]{0.49\textwidth}
    \includegraphics[width=\textwidth]{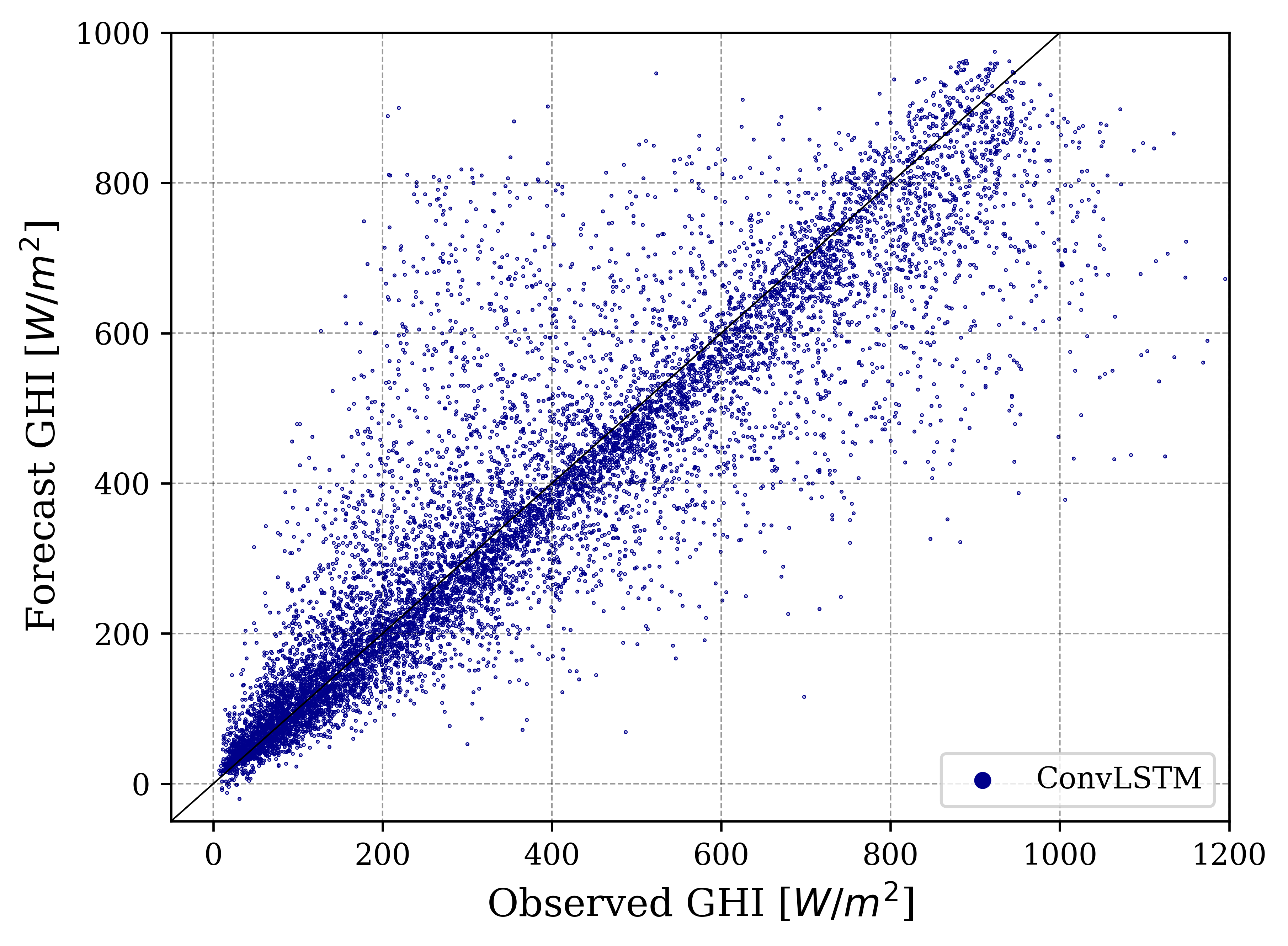}
    %\caption{Long exposure}
    \label{fig:scatter_plot_convlstm}
  \end{minipage}
  
\caption{10 min ahead test forecasts of the four benchmarked models (L2 loss function) and SIRTA's observations.}
\label{fig:scatter_plots}
\end{figure}

Notably, they often seem to improve over the persistence model by avoiding the double error caused by the time shift under sparse cloud cover. A typical example of this, is shown in Figure~\ref{fig:time_series_benchmark_1}. The ground truth and the persistence model curves are in phase opposition until 10:50 leading to a very high cumulative quadratic error. The strategy of the DL models to mitigate such behaviour seems to be to predict an intermediate irradiance in the convexity of the persistence model curve, which efficiently decreases the value of the largest errors over the same time period. This overestimation of low irradiance levels and underestimation of high irradiance levels is also visible in Figure~\ref{fig:scatter_plots}.

Moreover, it is interesting to notice that, from 10:50 to 11:30, the models consistently forecast an irradiance lower by at least $200 \; \text{W}/\text{m}^2$ than the smart persistence predictions although it is visible from the camera that the sun remains in a cloud-free area until the time point depicted on the image panel B' in Figure~\ref{fig:time_series_benchmark_1}. In similar conditions, this behaviour is also visible in Figure~\ref{fig:time_series_benchmark_4} from  9:00 to 9:30, but the corresponding shift is much smaller, from $10$ to $100\; \text{W}/\text{m}^2$. Compared to a $\text{L}_1$ loss, optimising a model with a $\text{L}_2$ loss function leads to a smaller variability of the prediction. This lack of dispersion in the forecast compared to observations is visible in Figure~\ref{fig:scatter_plots}. All observations above 1000 W/$\text{m}^2$ are underestimated with the model predictions remaining below 900 W/$\text{m}^2$ for the CNN, LSTM and 3D-CNN models and below 1000 W/$\text{m}^2$ for the ConvLSTM model. This suggests underfitting. However, the majority of the predictions lie close to the diagonal indicating that underfitting is limited to specific situations such as very high irradiance levels.

%Theses extreme events represent less than 1\% of all observations, which might explain why they are poorly taken into account by the models.

\subsection{Intra-hour irradiance forecasting}

We examine in this section the performance of the models on varying time horizons. The architectures used for the 10 min ahead forecast are now trained on other time horizons without further hyperparameter tuning. One can see in Table~\ref{results_fs_table_2_30min}, for forecasting horizons larger than 2 min, the 3D-CNN and ConvLSTM models outperform the other models. This being said, it is not clear that they could improve over the LSTM model on the shorter time horizon, i.e. 2 min. One of the possible reasons could be the need of hyperparameter tuning for this specific task. Indeed, a closer look at the FS of the ConvLSTM model shows that it is performing worse, comparatively, than the other models as the forecast window deviates from the initial targeted window, i.e. 10 min.

\renewcommand{\arraystretch}{1.1}
\begin{table}[H]
%\vskip 0.15in
\caption{RMSE forecast skill based on the SPM for different time horizons ($L_2$ loss function).}
\label{results_fs_table_2_30min}
\begin{center}
\begin{small}
%\begin{sc}
\begin{tabular}{lccccccccc}
Models & 2 min & 6 min & 10 min & 20 min & 30 min \\
\hline
%CNN & $L_1$ & &  &  &  & 17.2 &  &  &  \\

CNN & 7.5 & 16.9 & 18.1 & 19.2 & 19.7 \\
\\
%LSTM & $L_1$ & &  &  &  &  &  &  &  \\
LSTM & \textbf{10.8} & 16.5 & 19.2 & 20.4 & 20.9 \\
\\
%3D-CNN & $L_1$ & &  &  &  & 18.0 &  &  & \\
3D-CNN & 10.4 & \textbf{17.2} & 19.7 & 21.2 & \textbf{22.4} \\
 \\
%ConvLSTM & $L_1$ & &  &  &  & 18.7 &  &  &  \\
ConvLSTM & 9.05 & 16.6 & \textbf{20.4} & \textbf{21.4} & 22.1 \\
\hline
\end{tabular}
%\end{sc}
\end{small}
\end{center}
\vskip -0.1in
\end{table}

\vspace{1\baselineskip}

\begin{figure}[ht]%[ht!]%[h!] 
\centering    
\includegraphics[width=0.9\textwidth]{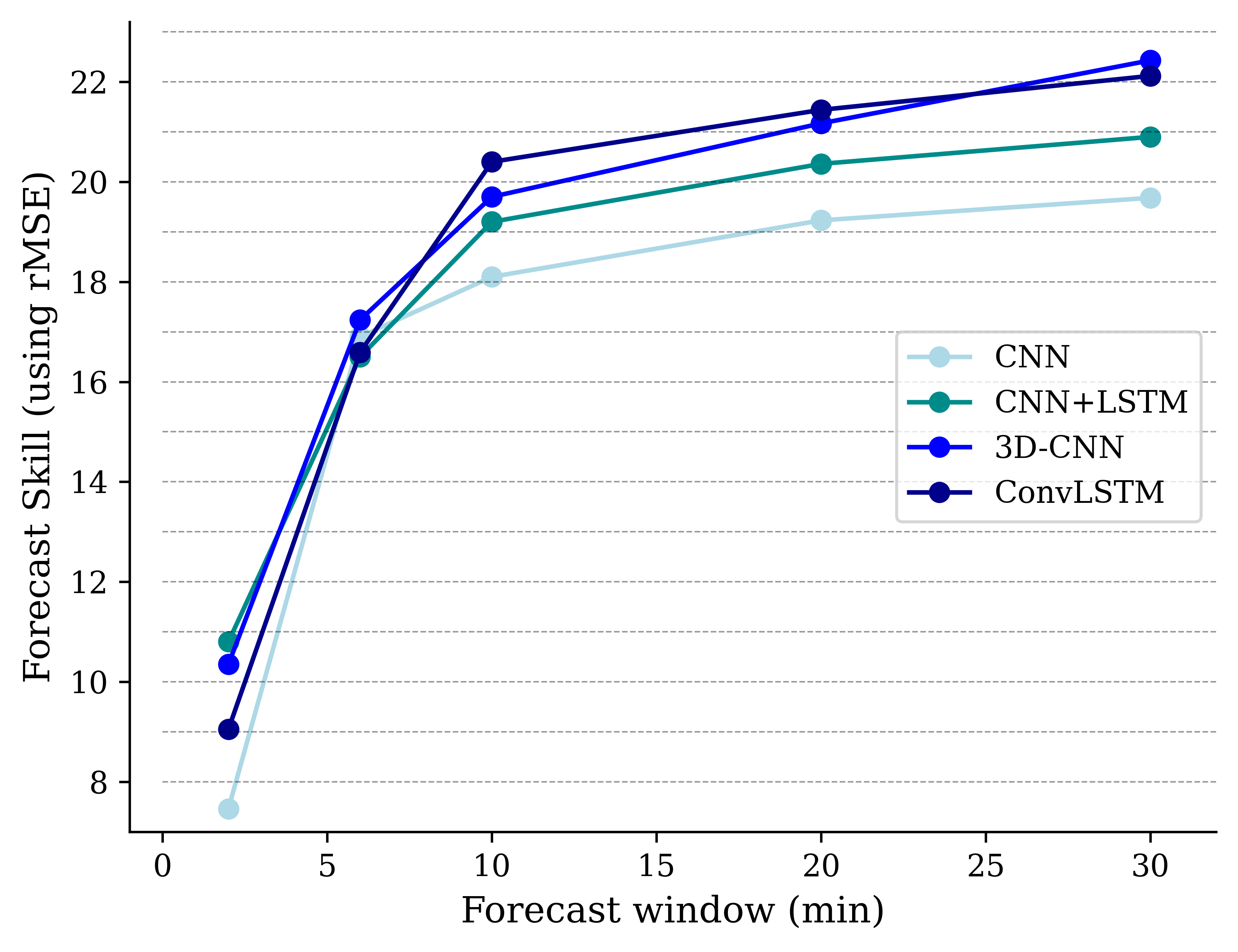}
\caption{Benchmark of the four model architectures for different forecast windows. A comparison with other deep learning methods based on skill scores reported in the literature is presented in~\ref{section:benchmark}.}
\label{fig:fs_rmse_benchmark}
\end{figure}

Experiments depicted in Figure~\ref{fig:fs_rmse_benchmark} show that models integrating temporal encoding of the data (LSTM), and in particular spatial temporal encoding (3D-CNN and ConvLSTM) of the sequence of images, have better generalisation abilities on the longest forecast windows, i.e. 20 and 30 min. However, the FS reached by all models range in the same window for shorter term forecasts. Interestingly, the ConvLSTM model sees the biggest FS loss relative to other models between the 10 and 2 min forecast windows. Presumably, this is caused by a greater dependency on the hyperparameter tuning for this architecture.

\subsection{Size of the training set}
\label{size_training_set}

\begin{figure}%[ht!]%[h!] 
\centering    
\includegraphics[width=0.9\textwidth]{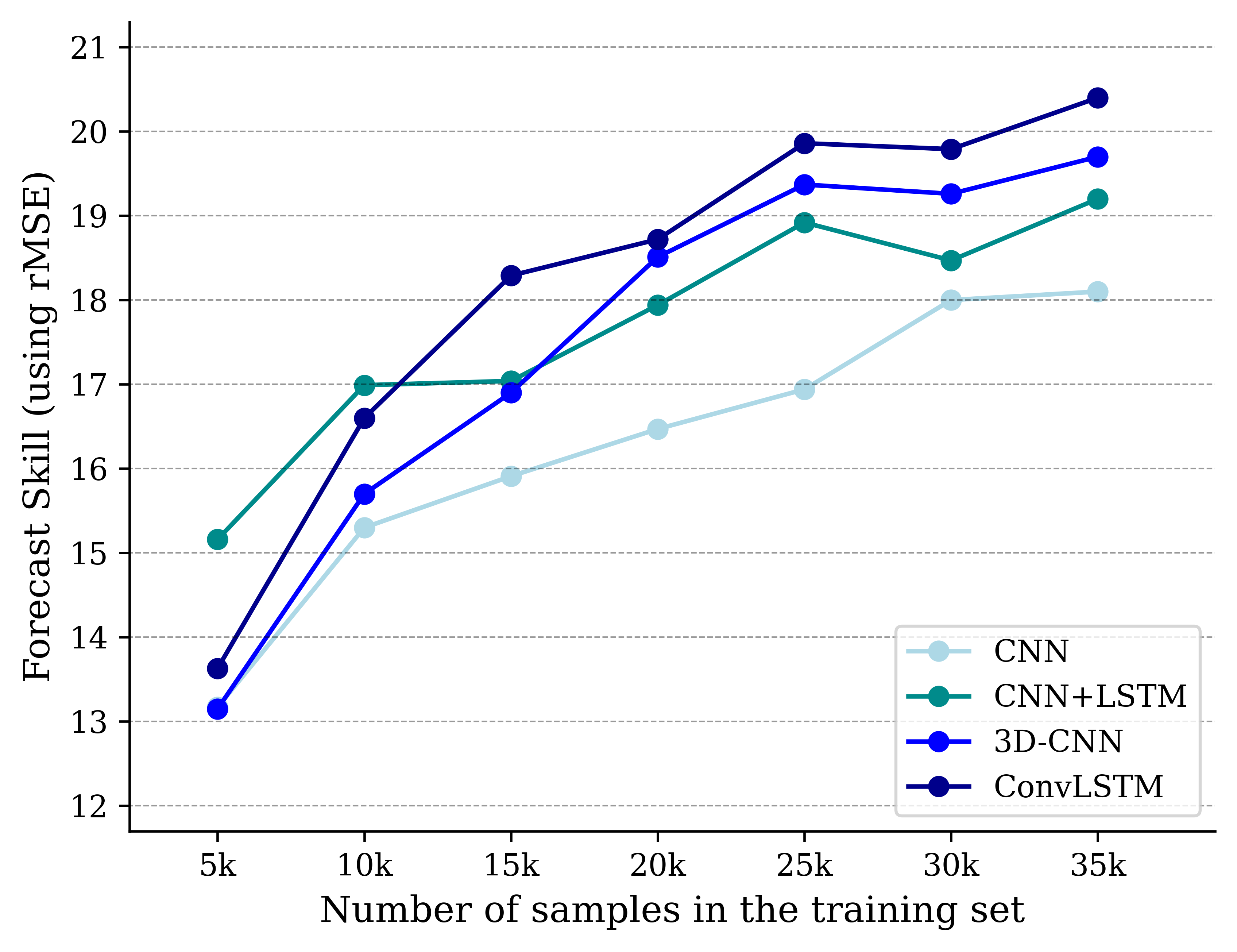}
\caption{Forecast skill for different size of the training set.}
\label{fig:fs_training_size}
\end{figure}

As a key aspect of data-driven approaches, the choice of the dataset has a significant impact on the performance of the DL models. In particular, the number of observations in the training set tends to be a determining variable with regards to generalisation abilities of the models. To highlight this point in the context of irradiance forecasting, all four models were trained on an increasing share of the total training set without further hyperparameter tuning. The corresponding forecast skill scores are reported in Figure~\ref{fig:fs_training_size}.

\vspace{1\baselineskip}

As shown here, increasing the size of the training set does play a critical role for this application, increasing the average FS from 14.1 for 5k samples to 19.6 for 35k samples. However, hyperparameters of the model were not specifically tuned for fewer training samples, which might explain some of the performance loss on small training sets. Yet, increasing the size of the training set from one to two years without additional hyperparameter tuning proved to increase the performance of the model further by about 10\% (10,000 samples from 2019 for both the validation and the test sets), confirming the initial insight (see Table~\ref{results_fs_1_vs_2_years_table}). Consequently, training the model on additional years of data or aggregated datasets from several locations may benefit the model's learning as much, if not more, than architectural tricks on these types of model.

\vspace{1\baselineskip}

\renewcommand{\arraystretch}{1.1}
\begin{table}
\caption{\label{results_fs_1_vs_2_years_table} RMSE forecast skill based on the SPM for different size of the training set. Second setting : 70,000 training samples (2017+2018), 10,000 validation samples (2019) and 10,000 test samples (2019).}
\begin{center}
\begin{small}
\begin{tabular}{lcccccccc}
Size of the training set & Loss & CNN & LSTM & 3D-CNN & ConvLSTM \\
\hline
\\
1 year (35k samples) & $\text{L}_1$ & 18.1 & 17.6 & 17.9 & 18.7 \\
& $\text{L}_2$ & 18.1 & 19.2 & 19.7 & 20.4 \\
\\
2 years (70k samples) & $\text{L}_1$ & 20.2 & 19.1 & 20.8 & 21.3 \\
& $\text{L}_2$ & \textbf{22.2} & \textbf{21.7} & \textbf{22.1} & \textbf{22.7} \\
\\
\hline
\end{tabular}
\end{small}
\end{center}
\vskip -0.1in
\end{table}

For the similar problem of rain forecasting from satellite imagery,  \cite{Sonderby2020} trained a ConvLSTM model on around 1.72 million training samples. Given that the size of available datasets for irradiance forecasting is more limited, this raises research topics such as transfer learning, data augmentation or dataset aggregation.

\section{Discussion}

Although DL seems to be a relevant approach to irradiance forecasting, this study shows that applying common architectures to solve this task gives mixed results. The different architectures prove to perform well for the optimisation problem of reaching a high forecast skill. However, such models are often missing the key objective of peak prediction for instance. As we can see in Figures~\ref{fig:time_series_benchmark_4} and~\ref{fig:time_series_benchmark_1}, the forecast methods are always late relative to the ground truth, which is quantified by TDM scores ranging from 0.34 to 0.64 (see Table~\ref{results_table}). Setting a different loss function penalising those behaviours~\citep{Guen2019} or carefully selecting the training samples may improve the value of the resulting forecasting models.

\vspace{1\baselineskip}

In addition, Section~\ref{size_training_set} points out that DL models do benefit significantly from increasing the amount of training data. This leads to the question of the use of data augmentation or transfer learning from a large dataset or an aggregation of many datasets from different solar facilities.

\vspace{1\baselineskip}

This being said, the general approach consisting of using a traditional model to forecast irradiance seems sub-optimal given the results presented in this work. The modelling of the temporal dynamics of the cloud cover might require a more sophisticated method compared to the one commonly used for the pairing of an image or sequence of images to its corresponding future irradiance. This could be achieved, for instance, using optical flow approaches~\citep{Black1992}, video prediction~\citep{Guen2020a, palettaECLIPSEEnvisioningCloud2021} or by isolating the region of interest in the current image given the past displacement of clouds~\citep{Quesada-Ruiz2014} to facilitate the learning. The delay in the predictions could be further penalised by incorporating this aspect of the forecast in the loss function~\citep{Guen2019}. Additionally, an adapted sampling strategy for training a model could improve its performance in specific conditions, such as high irradiance change events, which are critical but rare and thus underrepresented in a training set.

\section{Conclusion}
\label{conclusion}

Accurate forecasting of the irradiance and its largest variations is critical for the integration of solar facilities in the electrical grid. Besides statistical models, the current research in short-term forecasting involves the use of hemispherical cameras to capture real time cloudiness changes, which explains the variability of electricity production from solar panels. Given the growing number of large datasets, the deep learning framework appears to be a relevant candidate for this application. However, results obtained in this study reveal that there is still a progress margin. On one hand, common deep learning models have been shown to perform well for a given loss function, producing forecast skills of around 20\%. On the other hand, a closer look at other indicators such as the temporal distortion mix or the ramp score, shows that models tend to behave like a very smart persistence model, avoiding large errors at the cost of missing peaks and having regular time delays. Although the choice of the training set can impact the ability of the learning to generalise, it appears that the common approach of training a neural network to predict irradiance from a sequence of past images faces strong limitations. In addition to defining adapted sampling strategies and specific loss functions for deep irradiance forecasting, we suggest advancing the current irradiance forecasting framework by focusing on the dynamic aspects of the problem through video prediction or by locating the regions of interest as a preprocessing step. We hope that this will contribute to a shift of the deep learning approach to irradiance forecasting from reactive to anticipatory.

\section{Acknowledgements}

The authors acknowledge SIRTA for providing the sky images and irradiance measurements used in this study. We also thank Prof. Philippe Blanc for his guidance and valuable advice. We are grateful to Dr. Yves-Marie Saint-Drenan, Dr. Dmitry Slutskiy, Aleksandra Marconi and Anthony Hu for their technical assistance and valuable comments on the manuscript. This research was supported by ENGIE Lab CRIGEN, EPSRC and the University of Cambridge.

\bibliographystyle{elsarticle-harv} 
\bibliography{library.bib}

\begin{thebibliography}{56}
\expandafter\ifx\csname natexlab\endcsname\relax\def\natexlab#1{#1}\fi
\providecommand{\url}[1]{\texttt{#1}}
\providecommand{\href}[2]{#2}
\providecommand{\path}[1]{#1}
\providecommand{\DOIprefix}{doi:}
\providecommand{\ArXivprefix}{arXiv:}
\providecommand{\URLprefix}{URL: }
\providecommand{\Pubmedprefix}{pmid:}
\providecommand{\doi}[1]{\href{http://dx.doi.org/#1}{\path{#1}}}
\providecommand{\Pubmed}[1]{\href{pmid:#1}{\path{#1}}}
\providecommand{\bibinfo}[2]{#2}
\ifx\xfnm\relax \def\xfnm[#1]{\unskip,\space#1}\fi
%Type = Article
\bibitem[{Alonso and Batlles(2014)}]{Alonso2014}
\bibinfo{author}{Alonso, J.}, \bibinfo{author}{Batlles, F.J.},
  \bibinfo{year}{2014}.
\newblock \bibinfo{title}{{Short and medium-term cloudiness forecasting using
  remote sensing techniques and sky camera imagery}}.
\newblock \bibinfo{journal}{Energy} \bibinfo{volume}{73},
  \bibinfo{pages}{890--897}.
\newblock \URLprefix \url{http://dx.doi.org/10.1016/j.energy.2014.06.101},
  \DOIprefix\doi{10.1016/j.energy.2014.06.101}.
%Type = Article
\bibitem[{Bernecker et~al.(2014)Bernecker, Riess, Angelopoulou and
  Hornegger}]{Bernecker2014}
\bibinfo{author}{Bernecker, D.}, \bibinfo{author}{Riess, C.},
  \bibinfo{author}{Angelopoulou, E.}, \bibinfo{author}{Hornegger, J.},
  \bibinfo{year}{2014}.
\newblock \bibinfo{title}{{Continuous short-term irradiance forecasts using sky
  images}}.
\newblock \bibinfo{journal}{Solar Energy} \bibinfo{volume}{110},
  \bibinfo{pages}{303--315}.
\newblock \URLprefix \url{http://dx.doi.org/10.1016/j.solener.2014.09.005},
  \DOIprefix\doi{10.1016/j.solener.2014.09.005}.
%Type = Article
\bibitem[{Black(1992)}]{Black1992}
\bibinfo{author}{Black, M.J.}, \bibinfo{year}{1992}.
\newblock \bibinfo{title}{{Robust Incremental Optical Flow}}.
\newblock \bibinfo{journal}{Thesis} \bibinfo{volume}{1}, \bibinfo{pages}{280}.
%Type = Article
\bibitem[{Blanc et~al.(2011)Blanc, Gschwind, Lef{\`{e}}vre and
  Wald}]{Blanc2011}
\bibinfo{author}{Blanc, P.}, \bibinfo{author}{Gschwind, B.},
  \bibinfo{author}{Lef{\`{e}}vre, M.}, \bibinfo{author}{Wald, L.},
  \bibinfo{year}{2011}.
\newblock \bibinfo{title}{{The HelioClim Project: Surface Solar Irradiance Data
  for Climate Applications}}.
\newblock \bibinfo{journal}{Remote Sensing} \bibinfo{volume}{3},
  \bibinfo{pages}{343--361}.
\newblock \URLprefix \url{http://www.mdpi.com/2072-4292/3/2/343},
  \DOIprefix\doi{10.3390/rs3020343}.
%Type = Article
\bibitem[{Blanc et~al.(2017)Blanc, Massip, Kazantzidis, Tzoumanikas, Kuhn,
  Wilbert, Sch{\"{u}}ler and Prahl}]{Blanc2017}
\bibinfo{author}{Blanc, P.}, \bibinfo{author}{Massip, P.},
  \bibinfo{author}{Kazantzidis, A.}, \bibinfo{author}{Tzoumanikas, P.},
  \bibinfo{author}{Kuhn, P.}, \bibinfo{author}{Wilbert, S.},
  \bibinfo{author}{Sch{\"{u}}ler, D.}, \bibinfo{author}{Prahl, C.},
  \bibinfo{year}{2017}.
\newblock \bibinfo{title}{{Short-term forecasting of high resolution local DNI
  maps with multiple fish-eye cameras in stereoscopic mode}}.
\newblock \bibinfo{journal}{AIP Conference Proceedings} \bibinfo{volume}{1850}.
\newblock \DOIprefix\doi{10.1063/1.4984512}.
%Type = Article
\bibitem[{Bone et~al.(2018)Bone, Pidgeon, Kearney and Veeraragavan}]{Bone2018}
\bibinfo{author}{Bone, V.}, \bibinfo{author}{Pidgeon, J.},
  \bibinfo{author}{Kearney, M.}, \bibinfo{author}{Veeraragavan, A.},
  \bibinfo{year}{2018}.
\newblock \bibinfo{title}{{Intra-hour direct normal irradiance forecasting
  through adaptive clear-sky modelling and cloud tracking}}.
\newblock \bibinfo{journal}{Solar Energy} \bibinfo{volume}{159},
  \bibinfo{pages}{852--867}.
\newblock \URLprefix \url{https://doi.org/10.1016/j.solener.2017.10.037},
  \DOIprefix\doi{10.1016/j.solener.2017.10.037}.
%Type = Article
\bibitem[{Brad and Letia(2002)}]{Brad2002}
\bibinfo{author}{Brad, R.}, \bibinfo{author}{Letia, I.A.},
  \bibinfo{year}{2002}.
\newblock \bibinfo{title}{{Cloud motion detection from infrared satellite
  images}}.
\newblock \bibinfo{journal}{Second International Conference on Image and
  Graphics} \bibinfo{volume}{4875}, \bibinfo{pages}{408}.
\newblock \DOIprefix\doi{10.1117/12.477174}.
%Type = Article
\bibitem[{Chow et~al.(2011)Chow, Urquhart, Lave, Dominguez, Kleissl, Shields
  and Washom}]{Chow2011}
\bibinfo{author}{Chow, C.W.}, \bibinfo{author}{Urquhart, B.},
  \bibinfo{author}{Lave, M.}, \bibinfo{author}{Dominguez, A.},
  \bibinfo{author}{Kleissl, J.}, \bibinfo{author}{Shields, J.},
  \bibinfo{author}{Washom, B.}, \bibinfo{year}{2011}.
\newblock \bibinfo{title}{{Intra-hour forecasting with a total sky imager at
  the UC San Diego solar energy testbed}}.
\newblock \bibinfo{journal}{Solar Energy} \bibinfo{volume}{85},
  \bibinfo{pages}{2881--2893}.
\newblock \URLprefix \url{http://dx.doi.org/10.1016/j.solener.2011.08.025},
  \DOIprefix\doi{10.1016/j.solener.2011.08.025}.
%Type = Article
\bibitem[{Chu et~al.(2013)Chu, Pedro and Coimbra}]{Chu2013a}
\bibinfo{author}{Chu, Y.}, \bibinfo{author}{Pedro, H.T.},
  \bibinfo{author}{Coimbra, C.F.}, \bibinfo{year}{2013}.
\newblock \bibinfo{title}{{Hybrid intra-hour DNI forecasts with sky image
  processing enhanced by stochastic learning}}.
\newblock \bibinfo{journal}{Solar Energy} \bibinfo{volume}{98},
  \bibinfo{pages}{592--603}.
\newblock \URLprefix \url{http://dx.doi.org/10.1016/j.solener.2013.10.020},
  \DOIprefix\doi{10.1016/j.solener.2013.10.020}.
%Type = Article
\bibitem[{Ela et~al.(2013)Ela, Diakov, Ibanez and Heaney}]{Ela2013}
\bibinfo{author}{Ela, E.}, \bibinfo{author}{Diakov, V.},
  \bibinfo{author}{Ibanez, E.}, \bibinfo{author}{Heaney, M.},
  \bibinfo{year}{2013}.
\newblock \bibinfo{title}{{Impacts of Variability and Uncertainty in Solar
  Photovoltaic Generation at Multiple Timescales}}.
\newblock \bibinfo{journal}{National Renewable Energy Laboratory} \URLprefix
  \url{http://www.nrel.gov/docs/fy13osti/58274.pdf},
  \DOIprefix\doi{NREL/TP-5500-58274}.
%Type = Article
\bibitem[{Feng and Zhang(2020)}]{Feng2020}
\bibinfo{author}{Feng, C.}, \bibinfo{author}{Zhang, J.}, \bibinfo{year}{2020}.
\newblock \bibinfo{title}{{SolarNet: A sky image-based deep convolutional
  neural network for intra-hour solar forecasting}}.
\newblock \bibinfo{journal}{Solar Energy} \bibinfo{volume}{204},
  \bibinfo{pages}{71--78}.
\newblock \URLprefix \url{https://doi.org/10.1016/j.solener.2020.03.083},
  \DOIprefix\doi{10.1016/j.solener.2020.03.083}.
%Type = Article
\bibitem[{Florita et~al.(2013)Florita, Hodge and Orwig}]{Florita2013}
\bibinfo{author}{Florita, A.}, \bibinfo{author}{Hodge, B.M.},
  \bibinfo{author}{Orwig, K.}, \bibinfo{year}{2013}.
\newblock \bibinfo{title}{{Identifying wind and solar ramping events}}.
\newblock \bibinfo{journal}{IEEE Green Technologies Conference} ,
  \bibinfo{pages}{147--152}\DOIprefix\doi{10.1109/GreenTech.2013.30}.
%Type = Article
\bibitem[{Fr{\'{i}}as-Paredes et~al.(2016)Fr{\'{i}}as-Paredes, Mallor,
  Le{\'{o}}n and Gast{\'{o}}n-Romeo}]{Frias-Paredes2016}
\bibinfo{author}{Fr{\'{i}}as-Paredes, L.}, \bibinfo{author}{Mallor, F.},
  \bibinfo{author}{Le{\'{o}}n, T.}, \bibinfo{author}{Gast{\'{o}}n-Romeo, M.},
  \bibinfo{year}{2016}.
\newblock \bibinfo{title}{{Introducing the Temporal Distortion Index to perform
  a bidimensional analysis of renewable energy forecast}}.
\newblock \bibinfo{journal}{Energy} \bibinfo{volume}{94},
  \bibinfo{pages}{180--194}.
\newblock \DOIprefix\doi{10.1016/j.energy.2015.10.093}.
%Type = Article
\bibitem[{Guen and Thome(2019)}]{Guen2019}
\bibinfo{author}{Guen, V.L.}, \bibinfo{author}{Thome, N.},
  \bibinfo{year}{2019}.
\newblock \bibinfo{title}{{Shape and Time Distortion Loss for Training Deep
  Time Series Forecasting Models}}.
\newblock \bibinfo{journal}{Conference on Neural Information Processing
  Systems} \URLprefix \url{http://arxiv.org/abs/1909.09020},
  \href{http://arxiv.org/abs/1909.09020}{{\tt arXiv:1909.09020}}.
%Type = Article
\bibitem[{Guen et~al.(2020)Guen, Thome, Edf and National}]{Guen2020a}
\bibinfo{author}{Guen, V.L.}, \bibinfo{author}{Thome, N.},
  \bibinfo{author}{Edf, R.}, \bibinfo{author}{National, C.},
  \bibinfo{year}{2020}.
\newblock \bibinfo{title}{{A Deep Physical Model for Solar Irradiance
  Forecasting with Fisheye Images}}.
\newblock \bibinfo{journal}{CVPR} , \bibinfo{pages}{1--4}.
%Type = Article
\bibitem[{Haeffelin(2005)}]{sirta}
\bibinfo{author}{Haeffelin, M.}, \bibinfo{year}{2005}.
\newblock \bibinfo{title}{{SIRTA, a ground-based atmospheric observatory for
  cloud and aerosol research}}.
\newblock \bibinfo{journal}{Geophysicae} \bibinfo{volume}{23},
  \bibinfo{pages}{253--275}.
%Type = Article
\bibitem[{He et~al.(2016)He, Zhang, Ren and Sun}]{He2016}
\bibinfo{author}{He, K.}, \bibinfo{author}{Zhang, X.}, \bibinfo{author}{Ren,
  S.}, \bibinfo{author}{Sun, J.}, \bibinfo{year}{2016}.
\newblock \bibinfo{title}{{Deep Residual Learning for Image Recognition
  Kaiming}}.
\newblock \bibinfo{journal}{Conference on Computer Vision and Pattern
  Recognition} \DOIprefix\doi{10.1002/chin.200650130}.
%Type = Article
\bibitem[{Hochreiter and Schmidhuber(1997)}]{Hochreiter1997}
\bibinfo{author}{Hochreiter, S.}, \bibinfo{author}{Schmidhuber, J.},
  \bibinfo{year}{1997}.
\newblock \bibinfo{title}{{Long Short-Term Memory}}.
\newblock \bibinfo{journal}{Neural Computation} \bibinfo{volume}{9},
  \bibinfo{pages}{1735--1780}.
\newblock \DOIprefix\doi{10.1162/neco.1997.9.8.1735}.
%Type = Inproceedings
\bibitem[{Huang et~al.(2013)Huang, Xu, Peng, Yoo, Yu, Huang and
  Qin}]{huangCloudMotionEstimation2013}
\bibinfo{author}{Huang, H.}, \bibinfo{author}{Xu, J.}, \bibinfo{author}{Peng,
  Z.}, \bibinfo{author}{Yoo, S.}, \bibinfo{author}{Yu, D.},
  \bibinfo{author}{Huang, D.}, \bibinfo{author}{Qin, H.}, \bibinfo{year}{2013}.
\newblock \bibinfo{title}{Cloud motion estimation for short term solar
  irradiation prediction}, in: \bibinfo{booktitle}{2013 {{IEEE International
  Conference}} on {{Smart Grid Communications}} ({{SmartGridComm}})}, pp.
  \bibinfo{pages}{696--701}.
\newblock \DOIprefix\doi{10.1109/SmartGridComm.2013.6688040}.
%Type = Article
\bibitem[{Kingma and Ba(2015)}]{Kingma2015}
\bibinfo{author}{Kingma, D.P.}, \bibinfo{author}{Ba, J.L.},
  \bibinfo{year}{2015}.
\newblock \bibinfo{title}{{Adam: A method for stochastic optimization}}.
\newblock \bibinfo{journal}{3rd International Conference on Learning
  Representations, ICLR 2015 - Conference Track Proceedings} ,
  \bibinfo{pages}{1--15}\href{http://arxiv.org/abs/1412.6980}{{\tt
  arXiv:1412.6980}}.
%Type = Article
\bibitem[{Kuhn(2019)}]{Kuhn2019}
\bibinfo{author}{Kuhn, P.}, \bibinfo{year}{2019}.
\newblock \bibinfo{title}{{Development and Benchmarking of Solar Nowcasting
  Systems Entwicklung und Vergleich solarer
  K{\"{u}}rzestfrist-Vorhersagesysteme}}.
\newblock \bibinfo{journal}{Thesis} .
%Type = Article
\bibitem[{Kuhn et~al.(2019)Kuhn, Nouri, Wilbert, Hanrieder, Prahl, Ramirez,
  Zarzalejo, Schmidt, Schmidt, Yasser, Heinemann, Tzoumanikas, Kazantzidis,
  Kleissl, Blanc and {Pitz-Paal}}]{kuhnDeterminationOptimalCamera2019}
\bibinfo{author}{Kuhn, P.}, \bibinfo{author}{Nouri, B.},
  \bibinfo{author}{Wilbert, S.}, \bibinfo{author}{Hanrieder, N.},
  \bibinfo{author}{Prahl, C.}, \bibinfo{author}{Ramirez, L.},
  \bibinfo{author}{Zarzalejo, L.}, \bibinfo{author}{Schmidt, T.},
  \bibinfo{author}{Schmidt, T.}, \bibinfo{author}{Yasser, Z.},
  \bibinfo{author}{Heinemann, D.}, \bibinfo{author}{Tzoumanikas, P.},
  \bibinfo{author}{Kazantzidis, A.}, \bibinfo{author}{Kleissl, J.},
  \bibinfo{author}{Blanc, P.}, \bibinfo{author}{{Pitz-Paal}, R.},
  \bibinfo{year}{2019}.
\newblock \bibinfo{title}{Determination of the optimal camera distance for
  cloud height measurements with two all-sky imagers}.
\newblock \bibinfo{journal}{Solar Energy} \bibinfo{volume}{179},
  \bibinfo{pages}{74--88}.
\newblock \DOIprefix\doi{10.1016/j.solener.2018.12.038}.
%Type = Article
\bibitem[{Kwon and Park(2019)}]{Kwon2019}
\bibinfo{author}{Kwon, Y.H.}, \bibinfo{author}{Park, M.G.},
  \bibinfo{year}{2019}.
\newblock \bibinfo{title}{{Predicting future frames using retrospective cycle
  gan}}.
\newblock \bibinfo{journal}{Proceedings of the IEEE Computer Society Conference
  on Computer Vision and Pattern Recognition} \bibinfo{volume}{2019-June},
  \bibinfo{pages}{1811--1820}.
\newblock \DOIprefix\doi{10.1109/CVPR.2019.00191}.
%Type = Article
\bibitem[{LeCun et~al.(1989)LeCun, Boser, Denker, Henderson, Howard, Hubbard
  and Jackel}]{LeCun1989}
\bibinfo{author}{LeCun, Y.}, \bibinfo{author}{Boser, B.},
  \bibinfo{author}{Denker, J.S.}, \bibinfo{author}{Henderson, D.},
  \bibinfo{author}{Howard, R.E.}, \bibinfo{author}{Hubbard, W.},
  \bibinfo{author}{Jackel, L.D.}, \bibinfo{year}{1989}.
\newblock \bibinfo{title}{{Backpropagation Applied to Handwritten Zip Code
  Recognition}}.
\newblock \bibinfo{journal}{Neural Computation}
  \DOIprefix\doi{10.1162/neco.1989.1.4.541}.
%Type = Article
\bibitem[{Marquez and Coimbra(2013)}]{Marquez2013}
\bibinfo{author}{Marquez, R.}, \bibinfo{author}{Coimbra, C.F.},
  \bibinfo{year}{2013}.
\newblock \bibinfo{title}{{Intra-hour DNI forecasting based on cloud tracking
  image analysis}}.
\newblock \bibinfo{journal}{Solar Energy} \bibinfo{volume}{91},
  \bibinfo{pages}{327--336}.
\newblock \URLprefix \url{http://dx.doi.org/10.1016/j.solener.2012.09.018},
  \DOIprefix\doi{10.1016/j.solener.2012.09.018}.
%Type = Article
\bibitem[{Nair and Hinton(2010)}]{Nair2010}
\bibinfo{author}{Nair, V.}, \bibinfo{author}{Hinton, G.E.},
  \bibinfo{year}{2010}.
\newblock \bibinfo{title}{{Rectified Linear Units Improve Restricted Boltzmann
  Machines Vinod}}.
\newblock \bibinfo{journal}{ICML} \DOIprefix\doi{10.1123/jab.2016-0355}.
%Type = Article
\bibitem[{Nou et~al.(2018)Nou, Chauvin, Eynard, Thil and Grieu}]{Nou2018}
\bibinfo{author}{Nou, J.}, \bibinfo{author}{Chauvin, R.},
  \bibinfo{author}{Eynard, J.}, \bibinfo{author}{Thil, S.},
  \bibinfo{author}{Grieu, S.}, \bibinfo{year}{2018}.
\newblock \bibinfo{title}{{Towards the intrahour forecasting of direct normal
  irradiance using sky-imaging data}}.
\newblock \bibinfo{journal}{Heliyon} \bibinfo{volume}{4}.
\newblock \URLprefix \url{https://doi.org/10.1016/j.heliyon.2018.e00598},
  \DOIprefix\doi{10.1016/j.heliyon.2018.e00598}.
%Type = Article
\bibitem[{Nouri et~al.(2019a)Nouri, Kuhn, Wilbert, Hanrieder, Prahl, Zarzalejo,
  Kazantzidis, Blanc and {Pitz-Paal}}]{nouriCloudHeightTracking2019a}
\bibinfo{author}{Nouri, B.}, \bibinfo{author}{Kuhn, P.},
  \bibinfo{author}{Wilbert, S.}, \bibinfo{author}{Hanrieder, N.},
  \bibinfo{author}{Prahl, C.}, \bibinfo{author}{Zarzalejo, L.},
  \bibinfo{author}{Kazantzidis, A.}, \bibinfo{author}{Blanc, P.},
  \bibinfo{author}{{Pitz-Paal}, R.}, \bibinfo{year}{2019}a.
\newblock \bibinfo{title}{Cloud height and tracking accuracy of three all sky
  imager systems for individual clouds}.
\newblock \bibinfo{journal}{Solar Energy} \bibinfo{volume}{177},
  \bibinfo{pages}{213--228}.
\newblock \DOIprefix\doi{10.1016/j.solener.2018.10.079}.
%Type = Article
\bibitem[{Nouri et~al.(2018)Nouri, Kuhn, Wilbert, Prahl, {Pitz-Paal}, Blanc,
  Schmidt, Yasser, Santigosa and Heineman}]{nouriNowcastingDNIMaps2018}
\bibinfo{author}{Nouri, B.}, \bibinfo{author}{Kuhn, P.},
  \bibinfo{author}{Wilbert, S.}, \bibinfo{author}{Prahl, C.},
  \bibinfo{author}{{Pitz-Paal}, R.}, \bibinfo{author}{Blanc, P.},
  \bibinfo{author}{Schmidt, T.}, \bibinfo{author}{Yasser, Z.},
  \bibinfo{author}{Santigosa, L.R.}, \bibinfo{author}{Heineman, D.},
  \bibinfo{year}{2018}.
\newblock \bibinfo{title}{Nowcasting of {{DNI}} maps for the solar field based
  on voxel carving and individual {{3D}} cloud objects from all sky images}.
\newblock \bibinfo{journal}{AIP Conference Proceedings} \bibinfo{volume}{2033},
  \bibinfo{pages}{190011}.
\newblock \DOIprefix\doi{10.1063/1.5067196}.
%Type = Article
\bibitem[{Nouri et~al.(2020)Nouri, Wilbert, Blum, Kuhn, Schmidt, Yasser,
  Schmidt, Zarzalejo, Lopes, Silva, {Schroedter-Homscheidt}, Kazantzidis,
  Raeder, Blanc and {Pitz-Paal}}]{nouriEvaluationAllSky2020}
\bibinfo{author}{Nouri, B.}, \bibinfo{author}{Wilbert, S.},
  \bibinfo{author}{Blum, N.}, \bibinfo{author}{Kuhn, P.},
  \bibinfo{author}{Schmidt, T.}, \bibinfo{author}{Yasser, Z.},
  \bibinfo{author}{Schmidt, T.}, \bibinfo{author}{Zarzalejo, L.F.},
  \bibinfo{author}{Lopes, F.M.}, \bibinfo{author}{Silva, H.G.},
  \bibinfo{author}{{Schroedter-Homscheidt}, M.}, \bibinfo{author}{Kazantzidis,
  A.}, \bibinfo{author}{Raeder, C.}, \bibinfo{author}{Blanc, P.},
  \bibinfo{author}{{Pitz-Paal}, R.}, \bibinfo{year}{2020}.
\newblock \bibinfo{title}{Evaluation of an all sky imager based nowcasting
  system for distinct conditions and five sites}.
\newblock \bibinfo{journal}{AIP Conference Proceedings} \bibinfo{volume}{2303},
  \bibinfo{pages}{180006}.
\newblock \DOIprefix\doi{10.1063/5.0028670}.
%Type = Article
\bibitem[{Nouri et~al.(2019b)Nouri, Wilbert, Segura, Kuhn, Hanrieder,
  Kazantzidis, Schmidt, Zarzalejo, Blanc and
  {Pitz-Paal}}]{nouriDeterminationCloudTransmittance2019}
\bibinfo{author}{Nouri, B.}, \bibinfo{author}{Wilbert, S.},
  \bibinfo{author}{Segura, L.}, \bibinfo{author}{Kuhn, P.},
  \bibinfo{author}{Hanrieder, N.}, \bibinfo{author}{Kazantzidis, A.},
  \bibinfo{author}{Schmidt, T.}, \bibinfo{author}{Zarzalejo, L.},
  \bibinfo{author}{Blanc, P.}, \bibinfo{author}{{Pitz-Paal}, R.},
  \bibinfo{year}{2019}b.
\newblock \bibinfo{title}{Determination of cloud transmittance for all sky
  imager based solar nowcasting}.
\newblock \bibinfo{journal}{Solar Energy} \bibinfo{volume}{181},
  \bibinfo{pages}{251--263}.
\newblock \DOIprefix\doi{10.1016/j.solener.2019.02.004}.
%Type = Article
\bibitem[{Paletta et~al.(2021)Paletta, Hu, Arbod and
  Lasenby}]{palettaECLIPSEEnvisioningCloud2021}
\bibinfo{author}{Paletta, Q.}, \bibinfo{author}{Hu, A.},
  \bibinfo{author}{Arbod, G.}, \bibinfo{author}{Lasenby, J.},
  \bibinfo{year}{2021}.
\newblock \bibinfo{title}{{{ECLIPSE}} : {{Envisioning Cloud Induced
  Perturbations}} in {{Solar Energy}}}.
\newblock \bibinfo{journal}{arXiv:2104.12419 [cs]}
  \href{http://arxiv.org/abs/2104.12419}{{\tt arXiv:2104.12419}}.
%Type = Article
\bibitem[{Paletta and Lasenby(2020)}]{palettaConvolutionalNeuralNetworks2020a}
\bibinfo{author}{Paletta, Q.}, \bibinfo{author}{Lasenby, J.},
  \bibinfo{year}{2020}.
\newblock \bibinfo{title}{Convolutional {{Neural Networks}} applied to sky
  images for short-term solar irradiance forecasting}.
\newblock \bibinfo{journal}{arXiv:2005.11246 [cs, eess]}
  \href{http://arxiv.org/abs/2005.11246}{{\tt arXiv:2005.11246}}.
%Type = Article
\bibitem[{Pedro et~al.(2019)Pedro, Larson and Coimbra}]{Pedro2019}
\bibinfo{author}{Pedro, H.T.}, \bibinfo{author}{Larson, D.P.},
  \bibinfo{author}{Coimbra, C.F.}, \bibinfo{year}{2019}.
\newblock \bibinfo{title}{{A comprehensive dataset for the accelerated
  development and benchmarking of solar forecasting methods}}.
\newblock \bibinfo{journal}{Journal of Renewable and Sustainable Energy}
  \bibinfo{volume}{11}.
\newblock \DOIprefix\doi{10.1063/1.5094494}.
%Type = Inproceedings
\bibitem[{Peng et~al.(2014)Peng, Yoo, Yu, Huang, Kalb and
  Heiser}]{peng3DCloudDetection2014}
\bibinfo{author}{Peng, Z.}, \bibinfo{author}{Yoo, S.}, \bibinfo{author}{Yu,
  D.}, \bibinfo{author}{Huang, D.}, \bibinfo{author}{Kalb, P.},
  \bibinfo{author}{Heiser, J.}, \bibinfo{year}{2014}.
\newblock \bibinfo{title}{{{3D}} cloud detection and tracking for solar
  forecast using multiple sky imagers}, in: \bibinfo{booktitle}{Proceedings of
  the 29th {{Annual ACM Symposium}} on {{Applied Computing}}},
  \bibinfo{publisher}{{Association for Computing Machinery}},
  \bibinfo{address}{{New York, NY, USA}}. pp. \bibinfo{pages}{512--517}.
\newblock \DOIprefix\doi{10.1145/2554850.2554913}.
%Type = Article
\bibitem[{Peng et~al.(2016)Peng, Yu, Huang, Heiser and
  Kalb}]{pengHybridApproachEstimate2016}
\bibinfo{author}{Peng, Z.}, \bibinfo{author}{Yu, D.}, \bibinfo{author}{Huang,
  D.}, \bibinfo{author}{Heiser, J.}, \bibinfo{author}{Kalb, P.},
  \bibinfo{year}{2016}.
\newblock \bibinfo{title}{A hybrid approach to estimate the complex motions of
  clouds in sky images}.
\newblock \bibinfo{journal}{Solar Energy} \bibinfo{volume}{138},
  \bibinfo{pages}{10--25}.
\newblock \DOIprefix\doi{10.1016/j.solener.2016.09.002}.
%Type = Article
\bibitem[{Peng et~al.(2015)Peng, Yu, Huang, Heiser, Yoo and
  Kalb}]{peng3DCloudDetection2015a}
\bibinfo{author}{Peng, Z.}, \bibinfo{author}{Yu, D.}, \bibinfo{author}{Huang,
  D.}, \bibinfo{author}{Heiser, J.}, \bibinfo{author}{Yoo, S.},
  \bibinfo{author}{Kalb, P.}, \bibinfo{year}{2015}.
\newblock \bibinfo{title}{{{3D}} cloud detection and tracking system for solar
  forecast using multiple sky imagers}.
\newblock \bibinfo{journal}{Solar Energy} \bibinfo{volume}{118},
  \bibinfo{pages}{496--519}.
\newblock \DOIprefix\doi{10.1016/j.solener.2015.05.037}.
%Type = Incollection
\bibitem[{Perez and Hoff(2013)}]{Perez2013}
\bibinfo{author}{Perez, R.}, \bibinfo{author}{Hoff, T.E.},
  \bibinfo{year}{2013}.
\newblock \bibinfo{title}{{SolarAnywhere Forecasting}}, in:
  \bibinfo{booktitle}{Solar Energy Forecasting and Resource Assessment}.
  \bibinfo{publisher}{Academic Press, Boston}. chapter~\bibinfo{chapter}{10},
  pp. \bibinfo{pages}{233--265}.
\newblock \DOIprefix\doi{10.1016/B978-0-12-397177-7.00010-3}.
%Type = Inproceedings
\bibitem[{Pothineni et~al.(2019)Pothineni, Oswald, Poland and
  Pollefeys}]{pothineni2019b}
\bibinfo{author}{Pothineni, D.}, \bibinfo{author}{Oswald, M.R.},
  \bibinfo{author}{Poland, J.}, \bibinfo{author}{Pollefeys, M.},
  \bibinfo{year}{2019}.
\newblock \bibinfo{title}{Kloudnet: Deep learning for sky image analysis and
  irradiance forecasting}, in: \bibinfo{booktitle}{Pattern Recognition},
  \bibinfo{publisher}{Springer International Publishing}. pp.
  \bibinfo{pages}{535--551}.
%Type = Article
\bibitem[{Quesada-Ruiz et~al.(2014)Quesada-Ruiz, Chu, Tovar-Pescador, Pedro and
  Coimbra}]{Quesada-Ruiz2014}
\bibinfo{author}{Quesada-Ruiz, S.}, \bibinfo{author}{Chu, Y.},
  \bibinfo{author}{Tovar-Pescador, J.}, \bibinfo{author}{Pedro, H.T.},
  \bibinfo{author}{Coimbra, C.F.}, \bibinfo{year}{2014}.
\newblock \bibinfo{title}{{Cloud-tracking methodology for intra-hour DNI
  forecasting}}.
\newblock \bibinfo{journal}{Solar Energy} \bibinfo{volume}{102},
  \bibinfo{pages}{267--275}.
\newblock \URLprefix \url{http://dx.doi.org/10.1016/j.solener.2014.01.030},
  \DOIprefix\doi{10.1016/j.solener.2014.01.030}.
%Type = Article
\bibitem[{Schroff et~al.(2015)Schroff, Kalenichenko and Philbin}]{Schroff2015a}
\bibinfo{author}{Schroff, F.}, \bibinfo{author}{Kalenichenko, D.},
  \bibinfo{author}{Philbin, J.}, \bibinfo{year}{2015}.
\newblock \bibinfo{title}{{FaceNet: A unified embedding for face recognition
  and clustering}}.
\newblock \bibinfo{journal}{Proceedings of the IEEE Computer Society Conference
  on Computer Vision and Pattern Recognition} \bibinfo{volume}{07-12-June},
  \bibinfo{pages}{815--823}.
\newblock \DOIprefix\doi{10.1109/CVPR.2015.7298682},
  \href{http://arxiv.org/abs/1503.03832}{{\tt arXiv:1503.03832}}.
%Type = Article
\bibitem[{Shi et~al.(2015)Shi, Chen and Wang}]{Shi2015}
\bibinfo{author}{Shi, X.}, \bibinfo{author}{Chen, Z.}, \bibinfo{author}{Wang,
  H.}, \bibinfo{year}{2015}.
\newblock \bibinfo{title}{{Convolutional LSTM Network}}.
\newblock \bibinfo{journal}{Nips} , \bibinfo{pages}{2--3}\DOIprefix\doi{[]},
  \href{http://arxiv.org/abs/1506.04214}{{\tt arXiv:1506.04214}}.
%Type = Article
\bibitem[{Siddiqui et~al.(2019)Siddiqui, Bharadwaj and
  Kalyanaraman}]{Siddiqui2019}
\bibinfo{author}{Siddiqui, T.A.}, \bibinfo{author}{Bharadwaj, S.},
  \bibinfo{author}{Kalyanaraman, S.}, \bibinfo{year}{2019}.
\newblock \bibinfo{title}{{A deep learning approach to solar-irradiance
  forecasting in sky-videos}}.
\newblock \bibinfo{journal}{Proceedings - 2019 IEEE Winter Conference on
  Applications of Computer Vision, WACV 2019} ,
  \bibinfo{pages}{2166--2174}\DOIprefix\doi{10.1109/WACV.2019.00234},
  \href{http://arxiv.org/abs/1901.04881}{{\tt arXiv:1901.04881}}.
%Type = Article
\bibitem[{S{\o}nderby et~al.(2020)S{\o}nderby, Espeholt, Heek, Dehghani,
  Oliver, Salimans, Agrawal, Hickey and Kalchbrenner}]{Sonderby2020}
\bibinfo{author}{S{\o}nderby, C.K.}, \bibinfo{author}{Espeholt, L.},
  \bibinfo{author}{Heek, J.}, \bibinfo{author}{Dehghani, M.},
  \bibinfo{author}{Oliver, A.}, \bibinfo{author}{Salimans, T.},
  \bibinfo{author}{Agrawal, S.}, \bibinfo{author}{Hickey, J.},
  \bibinfo{author}{Kalchbrenner, N.}, \bibinfo{year}{2020}.
\newblock \bibinfo{title}{{MetNet: A Neural Weather Model for Precipitation
  Forecasting}}.
\newblock \bibinfo{journal}{ArXiv} , \bibinfo{pages}{1--17}\URLprefix
  \url{http://arxiv.org/abs/2003.12140},
  \href{http://arxiv.org/abs/2003.12140}{{\tt arXiv:2003.12140}}.
%Type = Article
\bibitem[{Sun et~al.(2018)Sun, Szucs and Brandt}]{Sun2018a}
\bibinfo{author}{Sun, Y.}, \bibinfo{author}{Szucs, G.},
  \bibinfo{author}{Brandt, A.R.}, \bibinfo{year}{2018}.
\newblock \bibinfo{title}{{Solar PV output prediction from video streams using
  convolutional neural networks}}.
\newblock \bibinfo{journal}{Energy and Environmental Science}
  \bibinfo{volume}{11}, \bibinfo{pages}{1811--1818}.
\newblock \DOIprefix\doi{10.1039/c7ee03420b}.
%Type = Article
\bibitem[{Vallance et~al.(2017)Vallance, Charbonnier, Paul, Dubost and
  Blanc}]{Vallance2017}
\bibinfo{author}{Vallance, L.}, \bibinfo{author}{Charbonnier, B.},
  \bibinfo{author}{Paul, N.}, \bibinfo{author}{Dubost, S.},
  \bibinfo{author}{Blanc, P.}, \bibinfo{year}{2017}.
\newblock \bibinfo{title}{{Towards a standardized procedure to assess solar
  forecast accuracy: A new ramp and time alignment metric}}.
\newblock \bibinfo{journal}{Solar Energy} \bibinfo{volume}{150},
  \bibinfo{pages}{408--422}.
\newblock \DOIprefix\doi{10.1016/j.solener.2017.04.064}.
%Type = Article
\bibitem[{Venugopal et~al.(2019)Venugopal, Sun and Brandt}]{Venugopal2019}
\bibinfo{author}{Venugopal, V.}, \bibinfo{author}{Sun, Y.},
  \bibinfo{author}{Brandt, A.R.}, \bibinfo{year}{2019}.
\newblock \bibinfo{title}{{Short-term solar PV forecasting using computer
  vision: The search for optimal CNN architectures for incorporating sky images
  and PV generation history}}.
\newblock \bibinfo{journal}{Journal of Renewable and Sustainable Energy}
  \bibinfo{volume}{11}.
\newblock \DOIprefix\doi{10.1063/1.5122796}.
%Type = Article
\bibitem[{Voyant et~al.(2017)Voyant, Notton, Kalogirou, Nivet, Paoli, Motte and
  Fouilloy}]{Voyant2017}
\bibinfo{author}{Voyant, C.}, \bibinfo{author}{Notton, G.},
  \bibinfo{author}{Kalogirou, S.}, \bibinfo{author}{Nivet, M.L.},
  \bibinfo{author}{Paoli, C.}, \bibinfo{author}{Motte, F.},
  \bibinfo{author}{Fouilloy, A.}, \bibinfo{year}{2017}.
\newblock \bibinfo{title}{{Machine learning methods for solar radiation
  forecasting: A review}}.
\newblock \bibinfo{journal}{Renewable Energy} \bibinfo{volume}{105},
  \bibinfo{pages}{569--582}.
\newblock \URLprefix \url{http://dx.doi.org/10.1016/j.renene.2016.12.095},
  \DOIprefix\doi{10.1016/j.renene.2016.12.095}.
%Type = Article
\bibitem[{Weinzaepfel et~al.(2013)Weinzaepfel, Revaud, Harchaoui and
  Schmid}]{Weinzaepfel2013a}
\bibinfo{author}{Weinzaepfel, P.}, \bibinfo{author}{Revaud, J.},
  \bibinfo{author}{Harchaoui, Z.}, \bibinfo{author}{Schmid, C.},
  \bibinfo{year}{2013}.
\newblock \bibinfo{title}{{DeepFlow: Large displacement optical flow with deep
  matching}}.
\newblock \bibinfo{journal}{Proceedings of the IEEE International Conference on
  Computer Vision} ,
  \bibinfo{pages}{1385--1392}\DOIprefix\doi{10.1109/ICCV.2013.175}.
%Type = Article
\bibitem[{Wen et~al.(2020)Wen, Du, Chen, Lim, Wen, Jiang and Xiang}]{Wen2020a}
\bibinfo{author}{Wen, H.}, \bibinfo{author}{Du, Y.}, \bibinfo{author}{Chen,
  X.}, \bibinfo{author}{Lim, E.}, \bibinfo{author}{Wen, H.},
  \bibinfo{author}{Jiang, L.}, \bibinfo{author}{Xiang, W.},
  \bibinfo{year}{2020}.
\newblock \bibinfo{title}{{Deep Learning-Based Multi-Step Solar Forecasting for
  PV Ramp-Rate Control Using Sky Images}}.
\newblock \bibinfo{journal}{IEEE Transactions on Industrial Informatics}
  \bibinfo{volume}{3203}, \bibinfo{pages}{1--1}.
\newblock \DOIprefix\doi{10.1109/tii.2020.2987916}.
%Type = Article
\bibitem[{Yadav and Chandel(2014)}]{Yadav2014}
\bibinfo{author}{Yadav, A.K.}, \bibinfo{author}{Chandel, S.S.},
  \bibinfo{year}{2014}.
\newblock \bibinfo{title}{{Solar radiation prediction using Artificial Neural
  Network techniques: A review}}.
\newblock \bibinfo{journal}{Renewable and Sustainable Energy Reviews}
  \bibinfo{volume}{33}, \bibinfo{pages}{772--781}.
\newblock \DOIprefix\doi{10.1016/j.rser.2013.08.055}.
%Type = Article
\bibitem[{Yang et~al.(2020)Yang, Alessandrini, Antonanzas, Antonanzas-Torres,
  Badescu, Beyer, Blaga, Boland, Bright, Coimbra, David, Frimane, Gueymard,
  Hong, Kay, Killinger, Kleissl, Lauret, Lorenz, van~der Meer, Paulescu, Perez,
  Perpi{\~{n}}{\'{a}}n-Lamigueiro, Peters, Reikard, Renn{\'{e}}, Saint-Drenan,
  Shuai, Urraca, Verbois, Vignola, Voyant and Zhang}]{Yang2020b}
\bibinfo{author}{Yang, D.}, \bibinfo{author}{Alessandrini, S.},
  \bibinfo{author}{Antonanzas, J.}, \bibinfo{author}{Antonanzas-Torres, F.},
  \bibinfo{author}{Badescu, V.}, \bibinfo{author}{Beyer, H.G.},
  \bibinfo{author}{Blaga, R.}, \bibinfo{author}{Boland, J.},
  \bibinfo{author}{Bright, J.M.}, \bibinfo{author}{Coimbra, C.F.},
  \bibinfo{author}{David, M.}, \bibinfo{author}{Frimane, {\^{A}}.},
  \bibinfo{author}{Gueymard, C.A.}, \bibinfo{author}{Hong, T.},
  \bibinfo{author}{Kay, M.J.}, \bibinfo{author}{Killinger, S.},
  \bibinfo{author}{Kleissl, J.}, \bibinfo{author}{Lauret, P.},
  \bibinfo{author}{Lorenz, E.}, \bibinfo{author}{van~der Meer, D.},
  \bibinfo{author}{Paulescu, M.}, \bibinfo{author}{Perez, R.},
  \bibinfo{author}{Perpi{\~{n}}{\'{a}}n-Lamigueiro, O.},
  \bibinfo{author}{Peters, I.M.}, \bibinfo{author}{Reikard, G.},
  \bibinfo{author}{Renn{\'{e}}, D.}, \bibinfo{author}{Saint-Drenan, Y.M.},
  \bibinfo{author}{Shuai, Y.}, \bibinfo{author}{Urraca, R.},
  \bibinfo{author}{Verbois, H.}, \bibinfo{author}{Vignola, F.},
  \bibinfo{author}{Voyant, C.}, \bibinfo{author}{Zhang, J.},
  \bibinfo{year}{2020}.
\newblock \bibinfo{title}{{Verification of deterministic solar forecasts}}.
\newblock \bibinfo{journal}{Solar Energy} , \bibinfo{pages}{1--18}\URLprefix
  \url{https://doi.org/10.1016/j.solener.2020.04.019},
  \DOIprefix\doi{10.1016/j.solener.2020.04.019}.
%Type = Article
\bibitem[{Yang et~al.(2018)Yang, Kleissl, Gueymard, Pedro and
  Coimbra}]{Yang2018}
\bibinfo{author}{Yang, D.}, \bibinfo{author}{Kleissl, J.},
  \bibinfo{author}{Gueymard, C.A.}, \bibinfo{author}{Pedro, H.T.},
  \bibinfo{author}{Coimbra, C.F.}, \bibinfo{year}{2018}.
\newblock \bibinfo{title}{{History and trends in solar irradiance and PV power
  forecasting: A preliminary assessment and review using text mining}}.
\newblock \bibinfo{journal}{Solar Energy} \bibinfo{volume}{168},
  \bibinfo{pages}{60--101}.
\newblock \DOIprefix\doi{10.1016/j.solener.2017.11.023}.
%Type = Article
\bibitem[{Zhang et~al.(2019)Zhang, Liwicki, Smith and Cipolla}]{Zhang2019}
\bibinfo{author}{Zhang, C.}, \bibinfo{author}{Liwicki, S.},
  \bibinfo{author}{Smith, W.}, \bibinfo{author}{Cipolla, R.},
  \bibinfo{year}{2019}.
\newblock \bibinfo{title}{{Orientation-aware semantic segmentation on
  icosahedron spheres}}.
\newblock \bibinfo{journal}{Proceedings of the IEEE International Conference on
  Computer Vision} \bibinfo{volume}{2019-Octob}, \bibinfo{pages}{3532--3540}.
\newblock \DOIprefix\doi{10.1109/ICCV.2019.00363},
  \href{http://arxiv.org/abs/1907.12849}{{\tt arXiv:1907.12849}}.
%Type = Article
\bibitem[{Zhang et~al.(2018)Zhang, Verschae, Nobuhara and Lalonde}]{Zhang2018}
\bibinfo{author}{Zhang, J.}, \bibinfo{author}{Verschae, R.},
  \bibinfo{author}{Nobuhara, S.}, \bibinfo{author}{Lalonde, J.F.},
  \bibinfo{year}{2018}.
\newblock \bibinfo{title}{{Deep photovoltaic nowcasting}}.
\newblock \bibinfo{journal}{Solar Energy} \bibinfo{volume}{176},
  \bibinfo{pages}{267--276}.
\newblock \URLprefix \url{https://doi.org/10.1016/j.solener.2018.10.024},
  \DOIprefix\doi{10.1016/j.solener.2018.10.024},
  \href{http://arxiv.org/abs/1810.06327}{{\tt arXiv:1810.06327}}.
%Type = Article
\bibitem[{Zhao et~al.(2019)Zhao, Wei, Wang, Zhu and Zhang}]{Zhao2019}
\bibinfo{author}{Zhao, X.}, \bibinfo{author}{Wei, H.}, \bibinfo{author}{Wang,
  H.}, \bibinfo{author}{Zhu, T.}, \bibinfo{author}{Zhang, K.},
  \bibinfo{year}{2019}.
\newblock \bibinfo{title}{{3D-CNN-based feature extraction of ground-based
  cloud images for direct normal irradiance prediction}}.
\newblock \bibinfo{journal}{Solar Energy} \bibinfo{volume}{181},
  \bibinfo{pages}{510--518}.
\newblock \URLprefix \url{https://doi.org/10.1016/j.solener.2019.01.096},
  \DOIprefix\doi{10.1016/j.solener.2019.01.096}.

\end{thebibliography}

\newpage

\appendix

\section{Dataset Balance}
\label{section:dataset_balance}

\begin{figure}[H]%[ht!]%[h!] 
\centering
\begin{minipage}[b]{0.49\textwidth}
    \includegraphics[width=\textwidth]{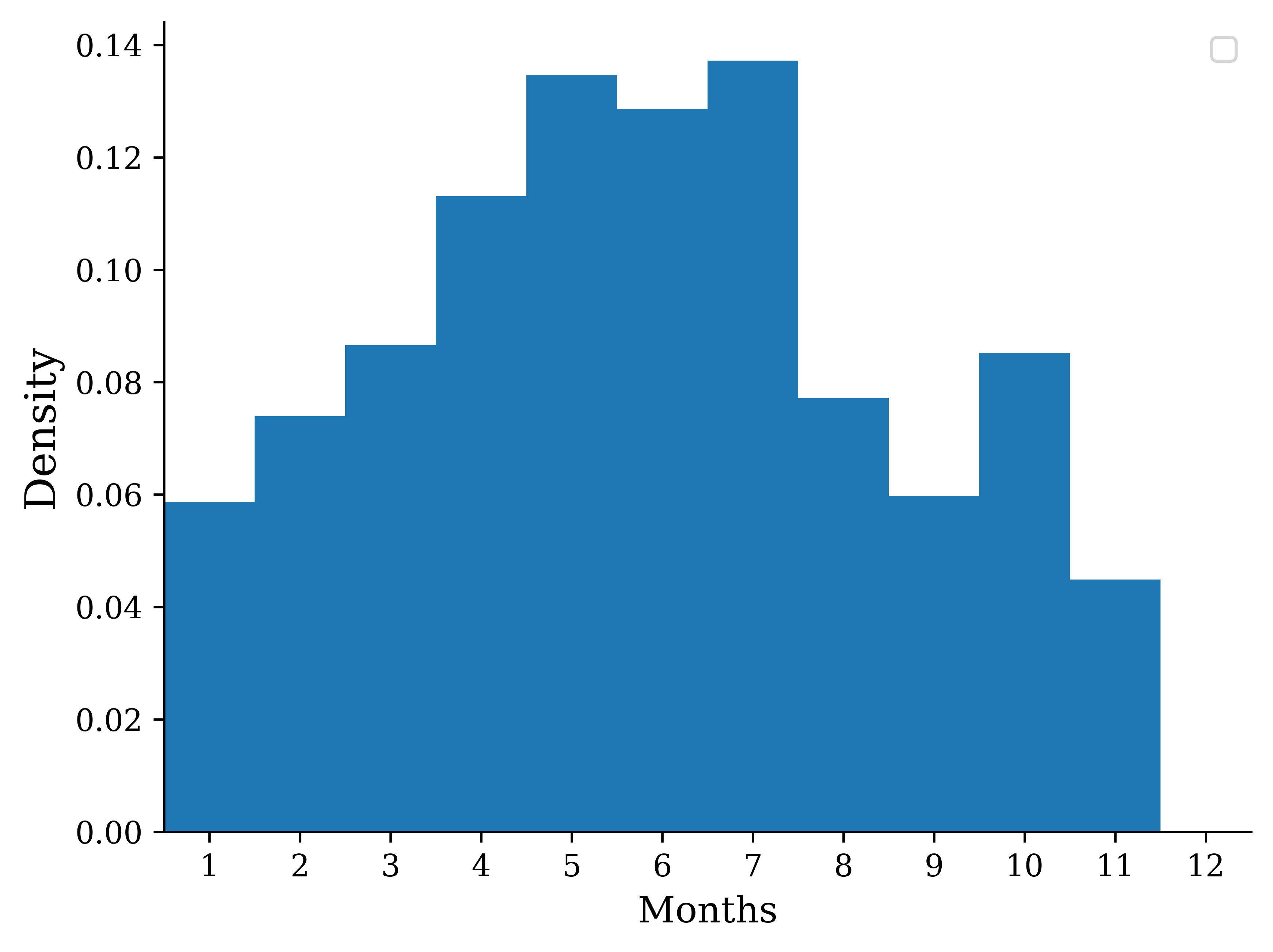}
    %\caption{Short exposure}
    \label{fig:hist_months_2017}
  \end{minipage} 
  %\quad
  \begin{minipage}[b]{0.49\textwidth}
    \includegraphics[width=\textwidth]{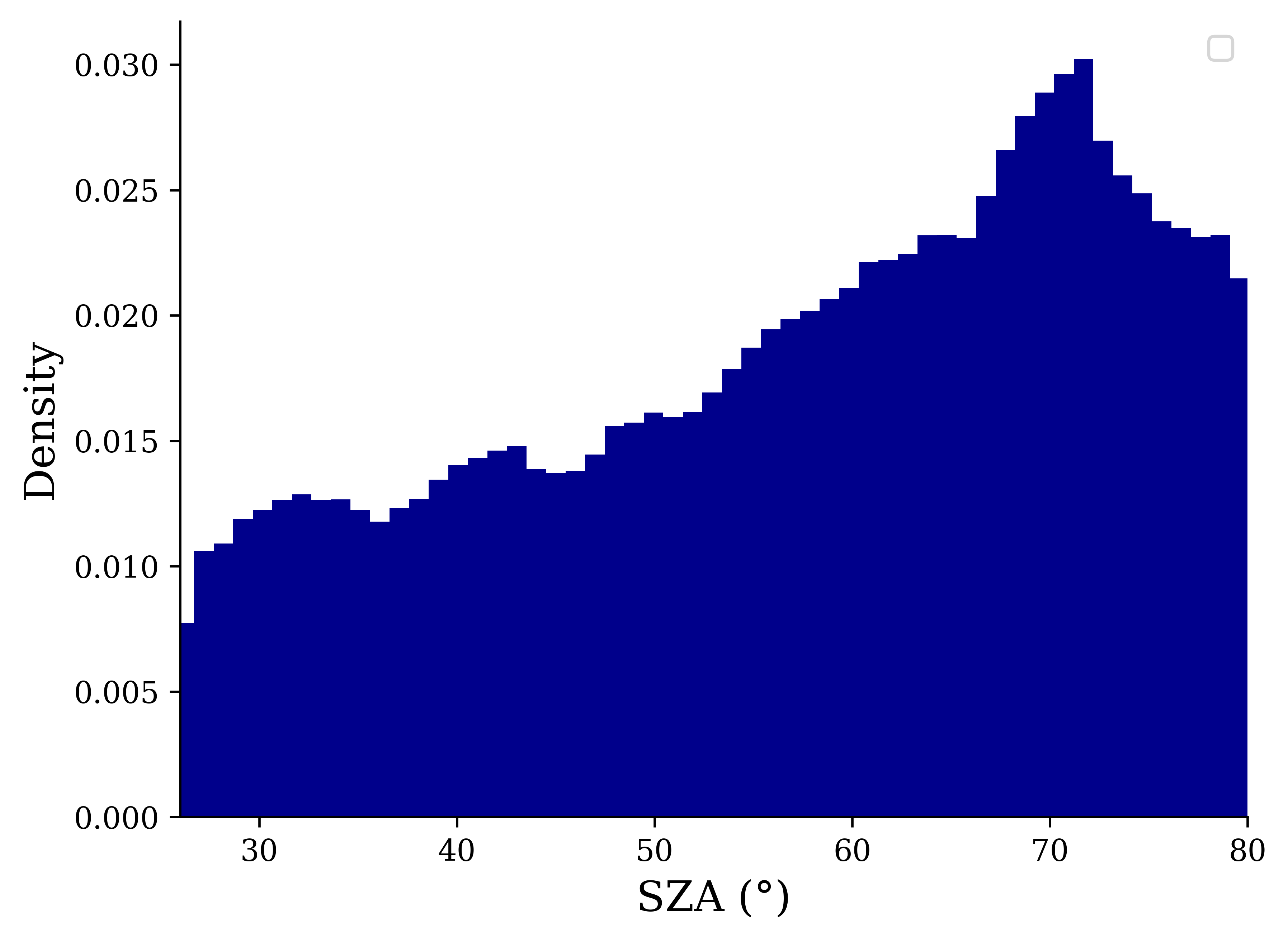}
    %\caption{Long exposure}
    \label{fig:hist_sza_2017}
  \end{minipage}
\caption{Distribution of samples in the training set (2017).}
\label{fig:hist_2017}
\end{figure}

\begin{figure}[H]%[ht!]%[h!] 
\centering
\begin{minipage}[b]{0.49\textwidth}
    \includegraphics[width=\textwidth]{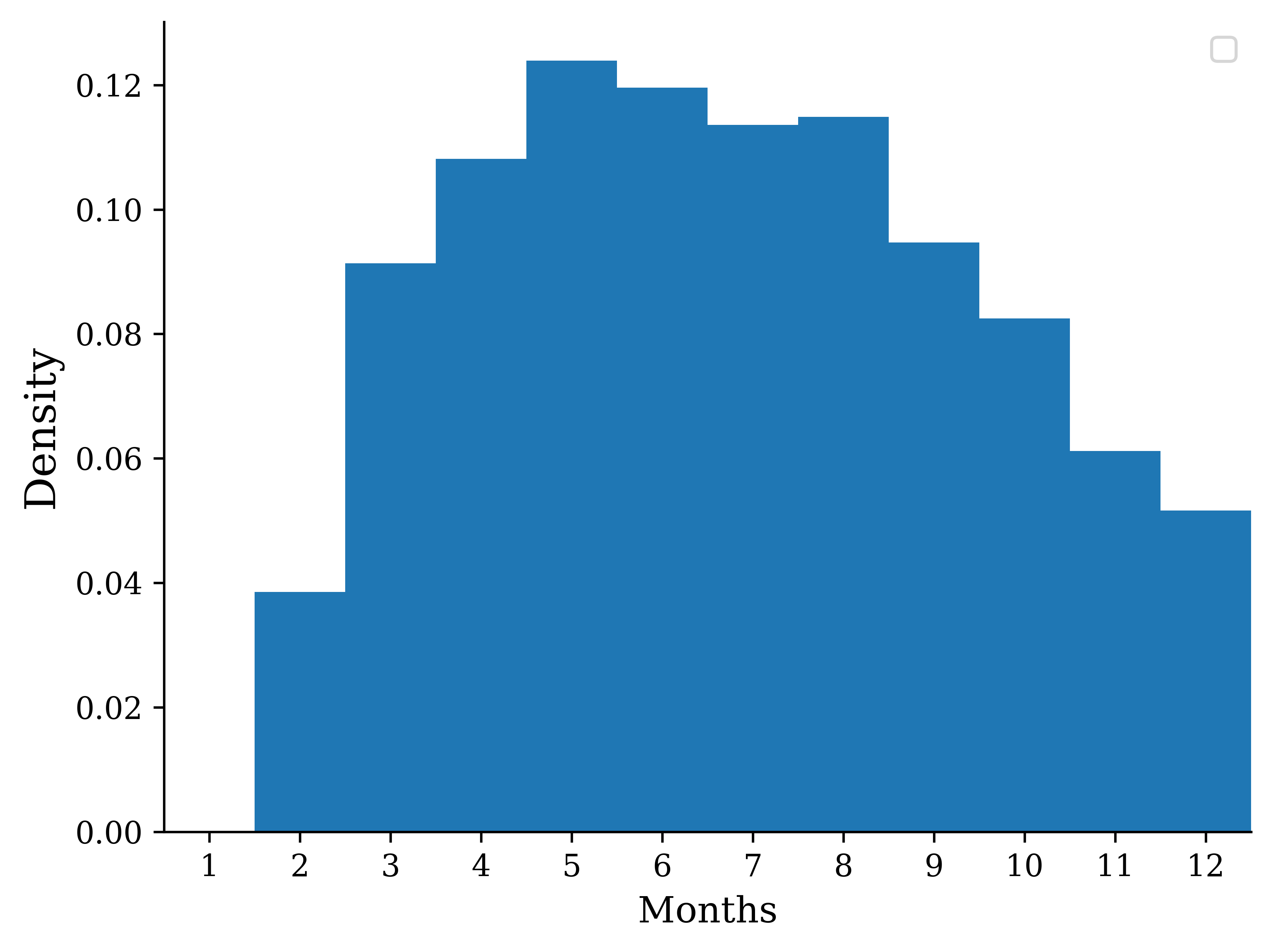}
    %\caption{Short exposure}
    \label{fig:hist_months_2018}
  \end{minipage} 
  %\quad
  \begin{minipage}[b]{0.49\textwidth}
    \includegraphics[width=\textwidth]{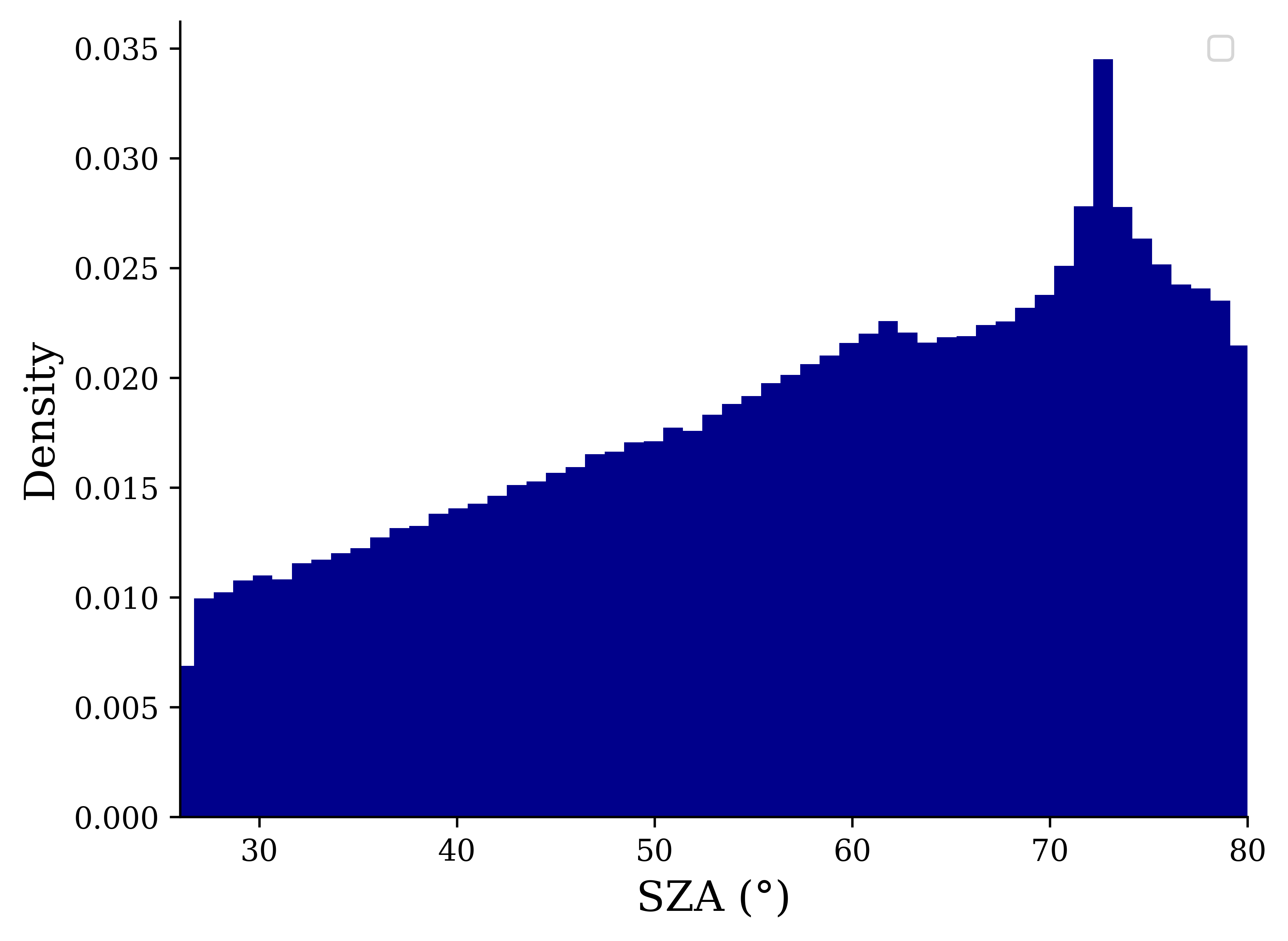}
    %\caption{Long exposure}
    \label{fig:hist_sza_2018}
  \end{minipage}
\caption{Distribution of samples in the validation set (2018).}
\label{fig:hist_2018}
\end{figure}

\begin{figure}[H]%[ht!]%[h!] 
\centering
\begin{minipage}[b]{0.49\textwidth}
    \includegraphics[width=\textwidth]{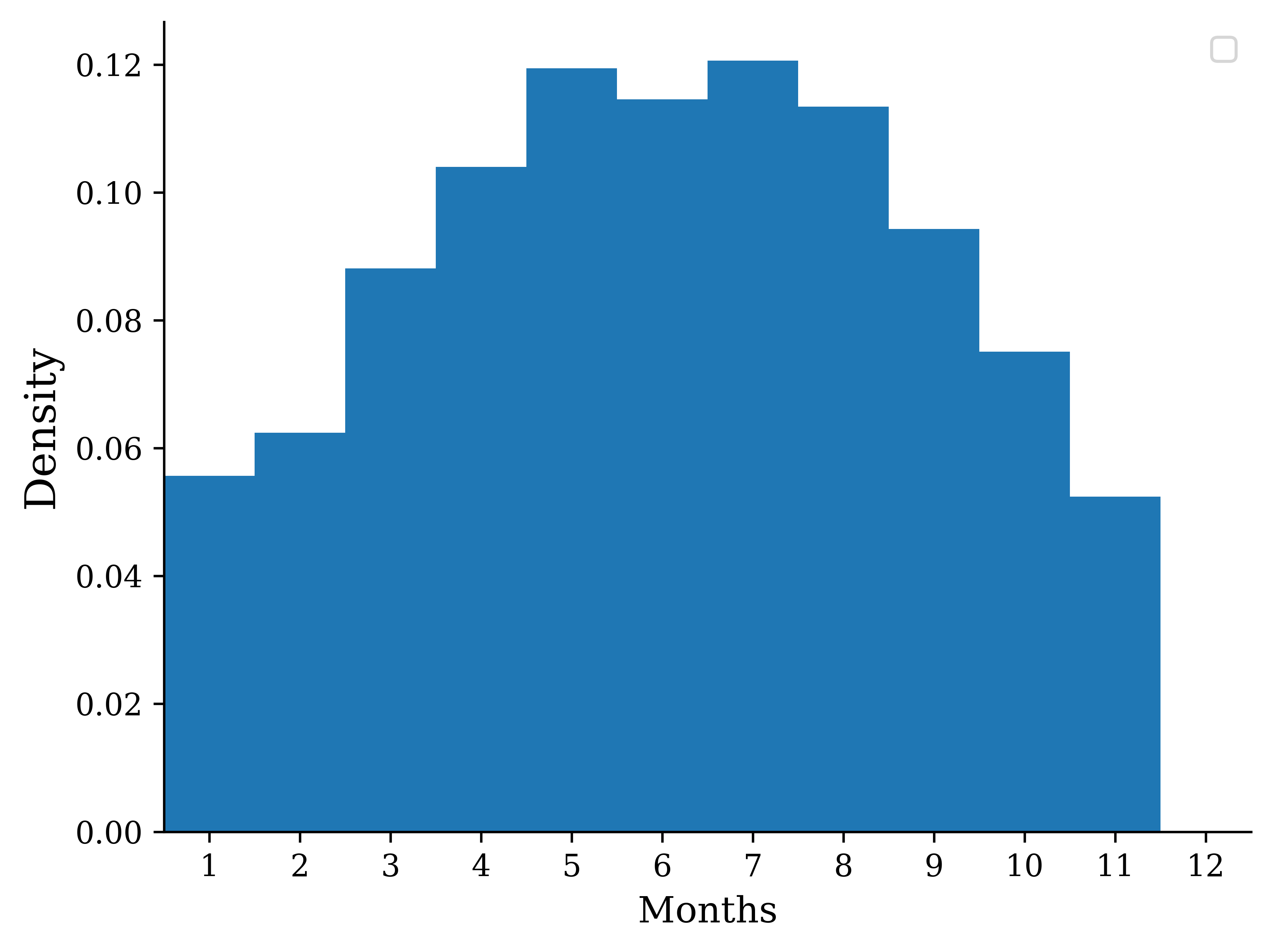}
    %\caption{Short exposure}
    \label{fig:hist_months_2019}
  \end{minipage} 
  %\quad
  \begin{minipage}[b]{0.49\textwidth}
    \includegraphics[width=\textwidth]{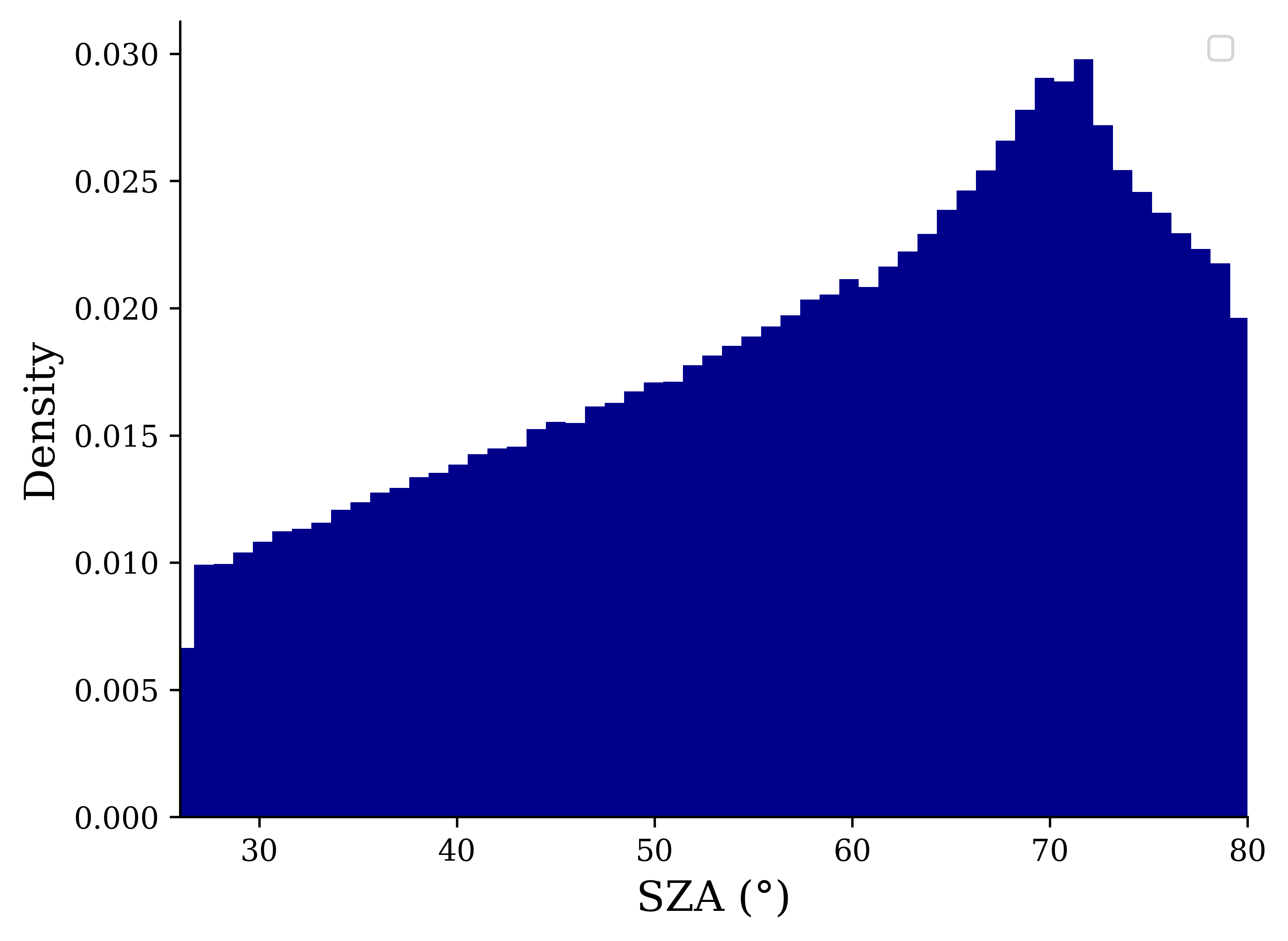}
    %\caption{Long exposure}
    \label{fig:hist_sza_2019}
  \end{minipage}
\caption{Distribution of samples in the test set (2019).}
\label{fig:hist_2019}
\end{figure}

\newpage

\section{Architectures}
\label{section:architectures}

\begin{figure}[H]%[ht!]%[h!] 
\centering    
\includegraphics[width=1\textwidth]{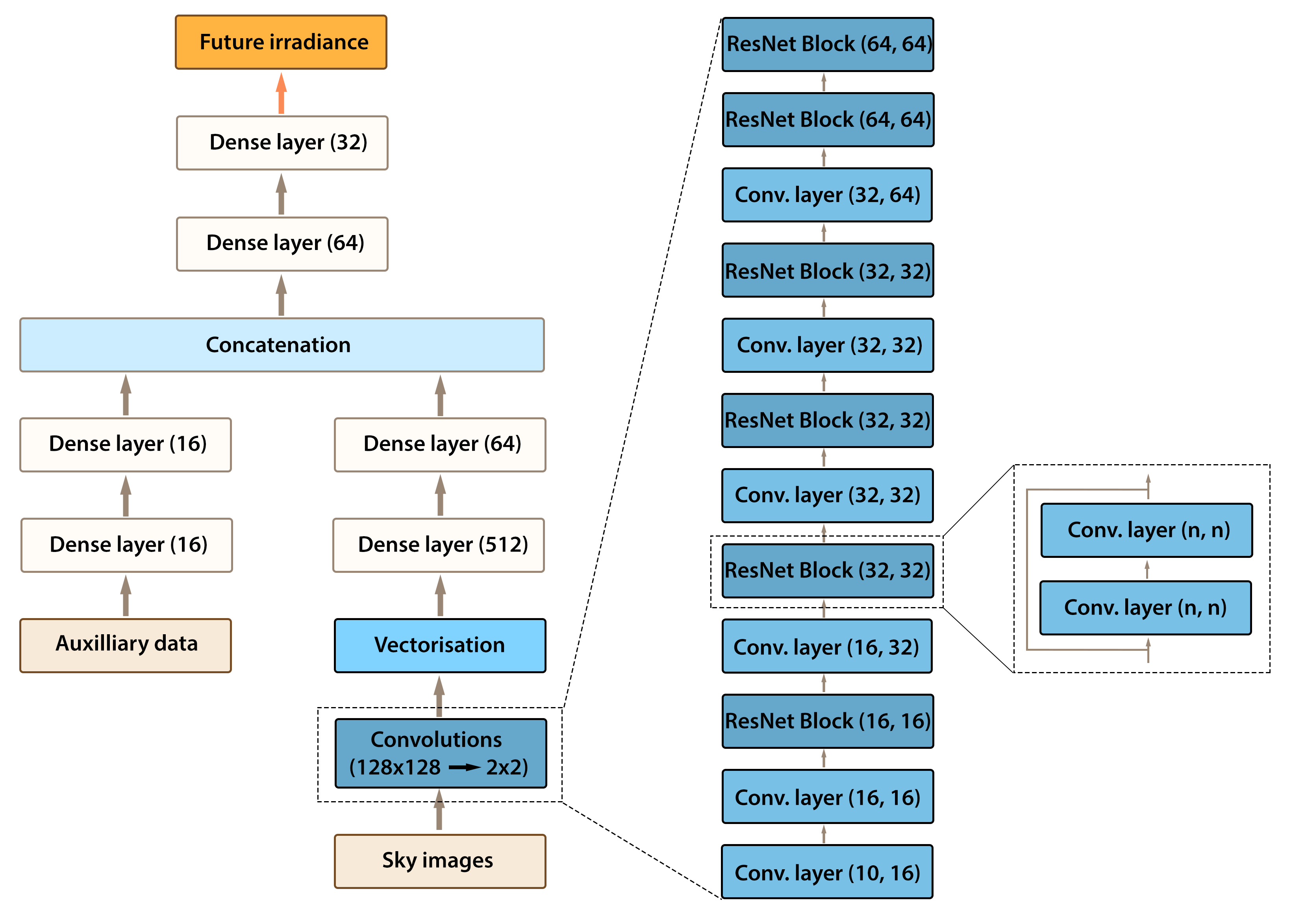}
\caption{CNN model detailed architecture.}
\label{fig:Figure_arch_CNN}
\end{figure}

\begin{figure}[H]%[ht!]%[h!] 
\centering    
\includegraphics[width=1\textwidth]{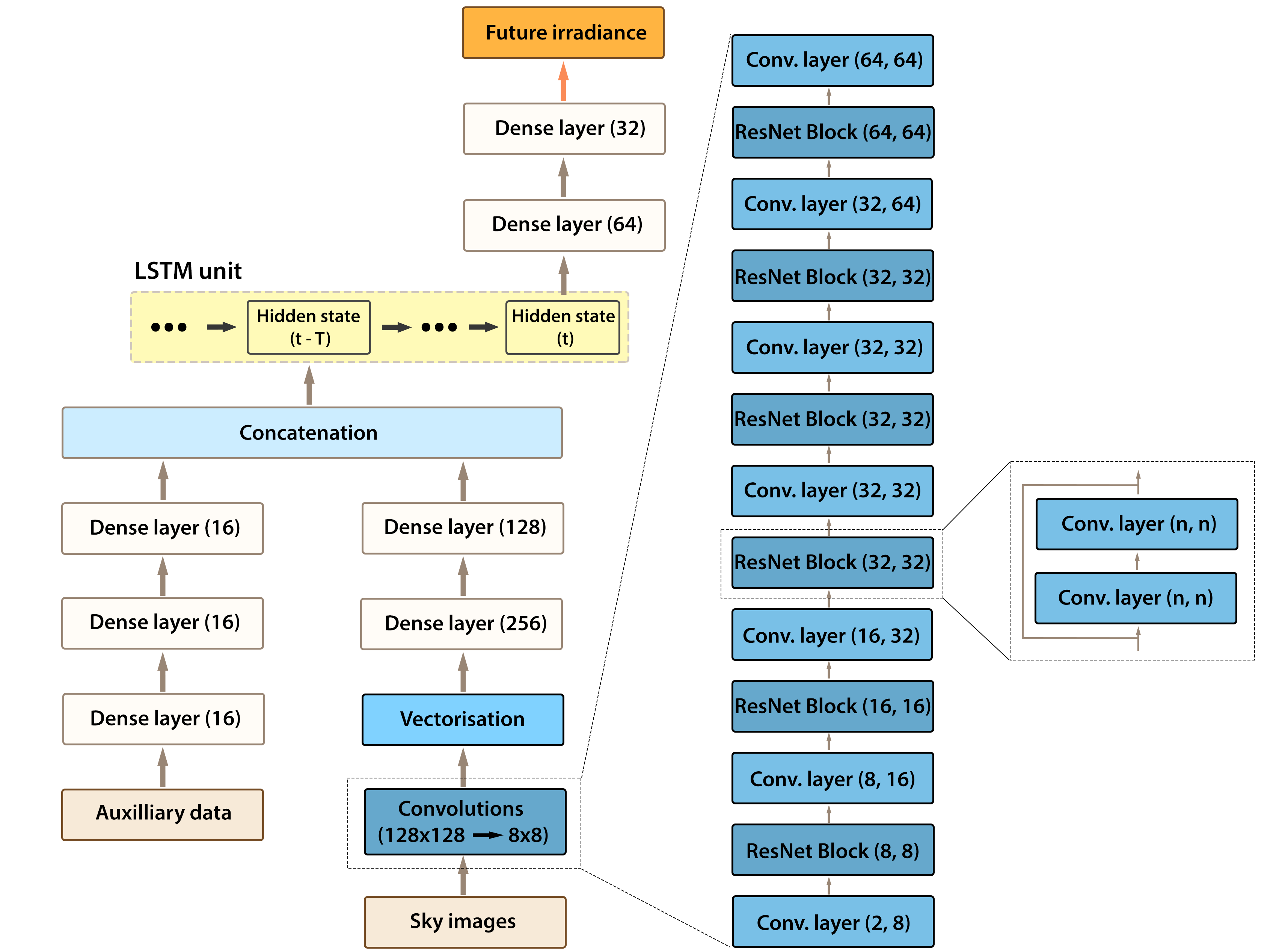}
\caption{LSTM model detailed architecture.}
\label{fig:Figure_arch_LSTM}
\end{figure}

\begin{figure}[H]%[ht!]%[h!] 
\centering    
\includegraphics[width=1\textwidth]{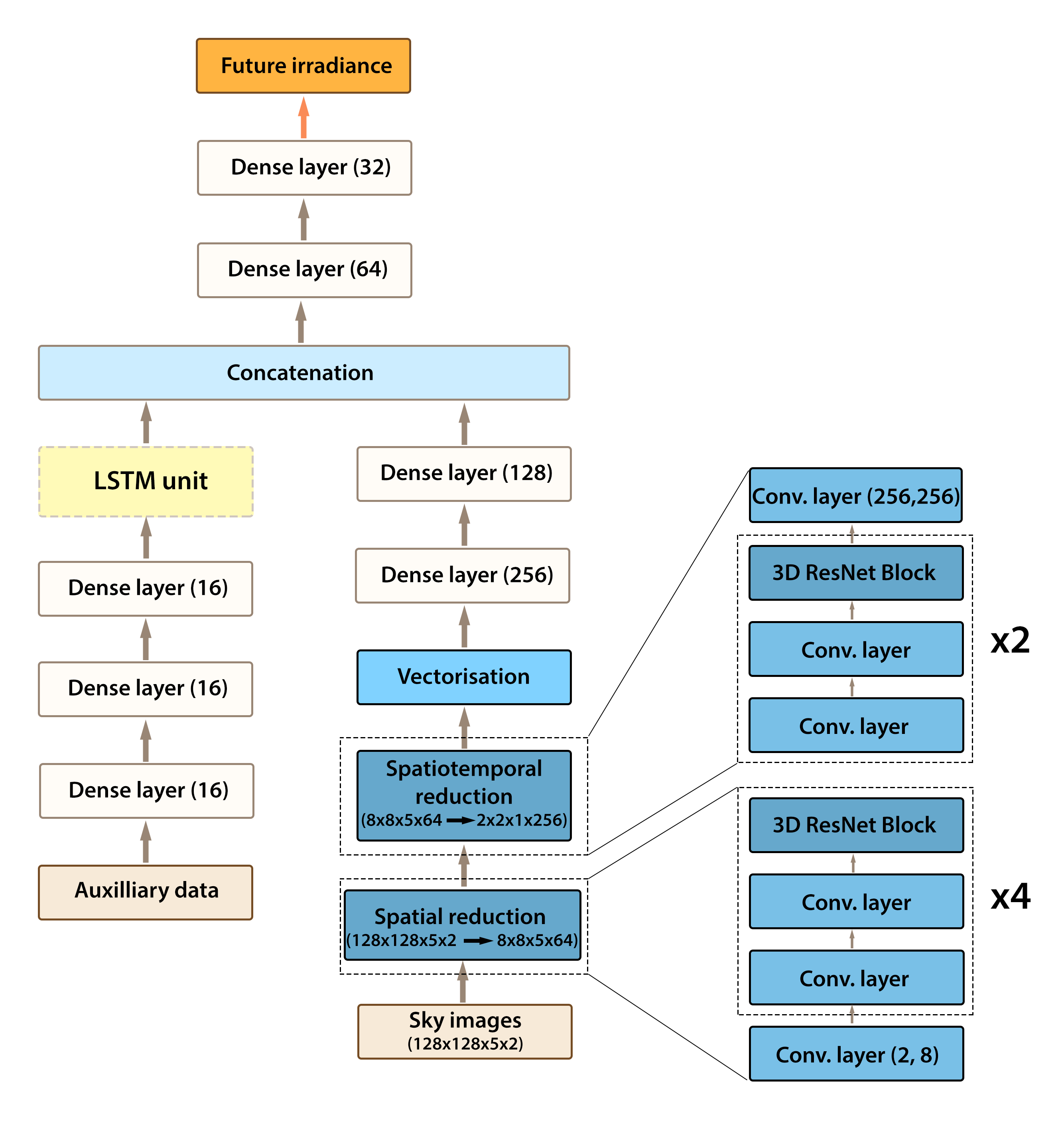}
\caption{3D-CNN model detailed architecture.}
\label{fig:Figure_arch_3D}
\end{figure}

\begin{figure}[H]%[ht!]%[h!] 
\centering    
\includegraphics[width=1\textwidth]{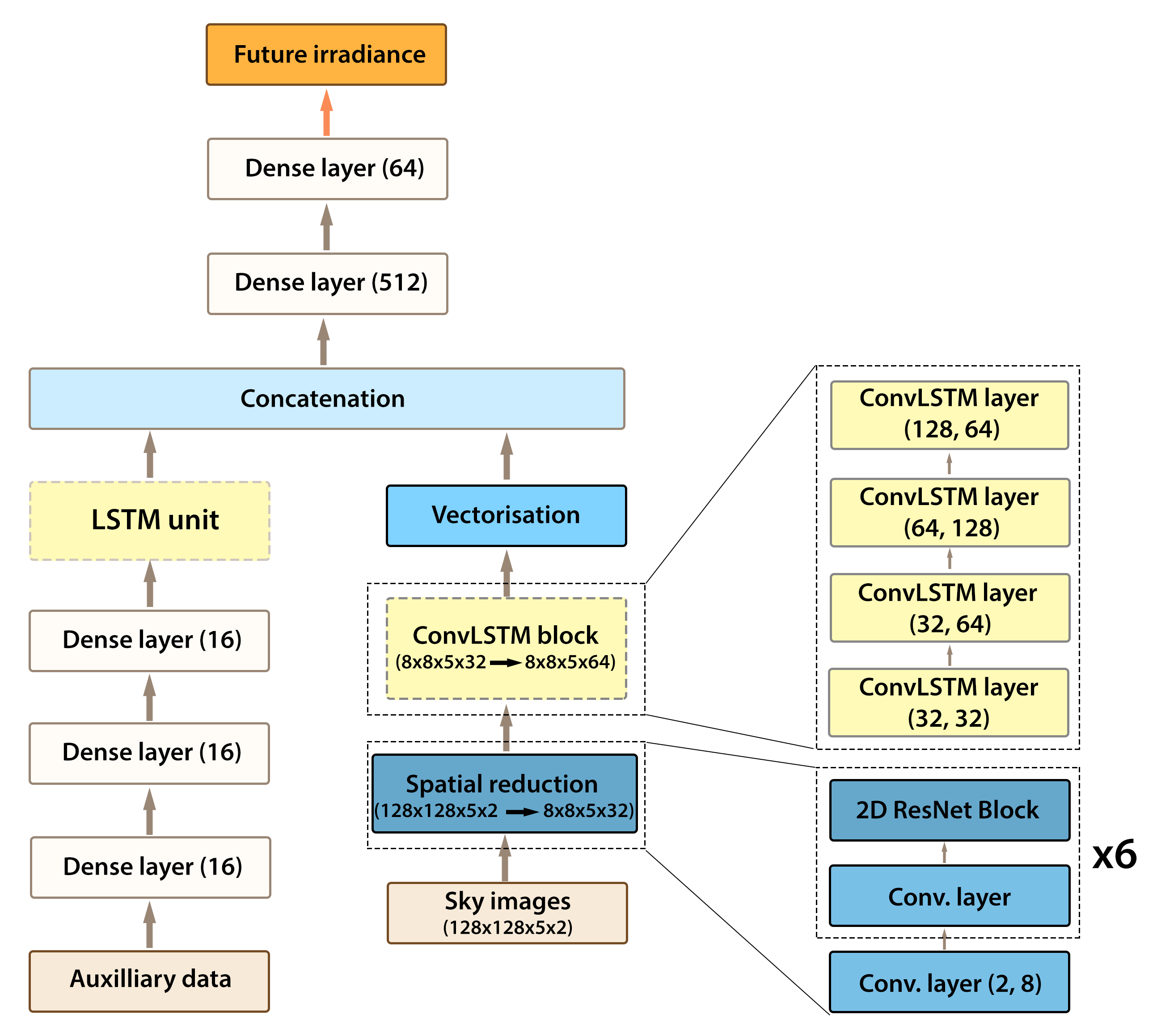}
\caption{ConvLSTM model detailed architecture.}
\label{fig:Figure_arch_ConvLSTM}
\end{figure}

\newpage

\section{Benchmark with other methods}
\label{section:benchmark}

\begin{figure}[ht]%[ht!]%[h!] 
\centering    
\includegraphics[width=1\textwidth]{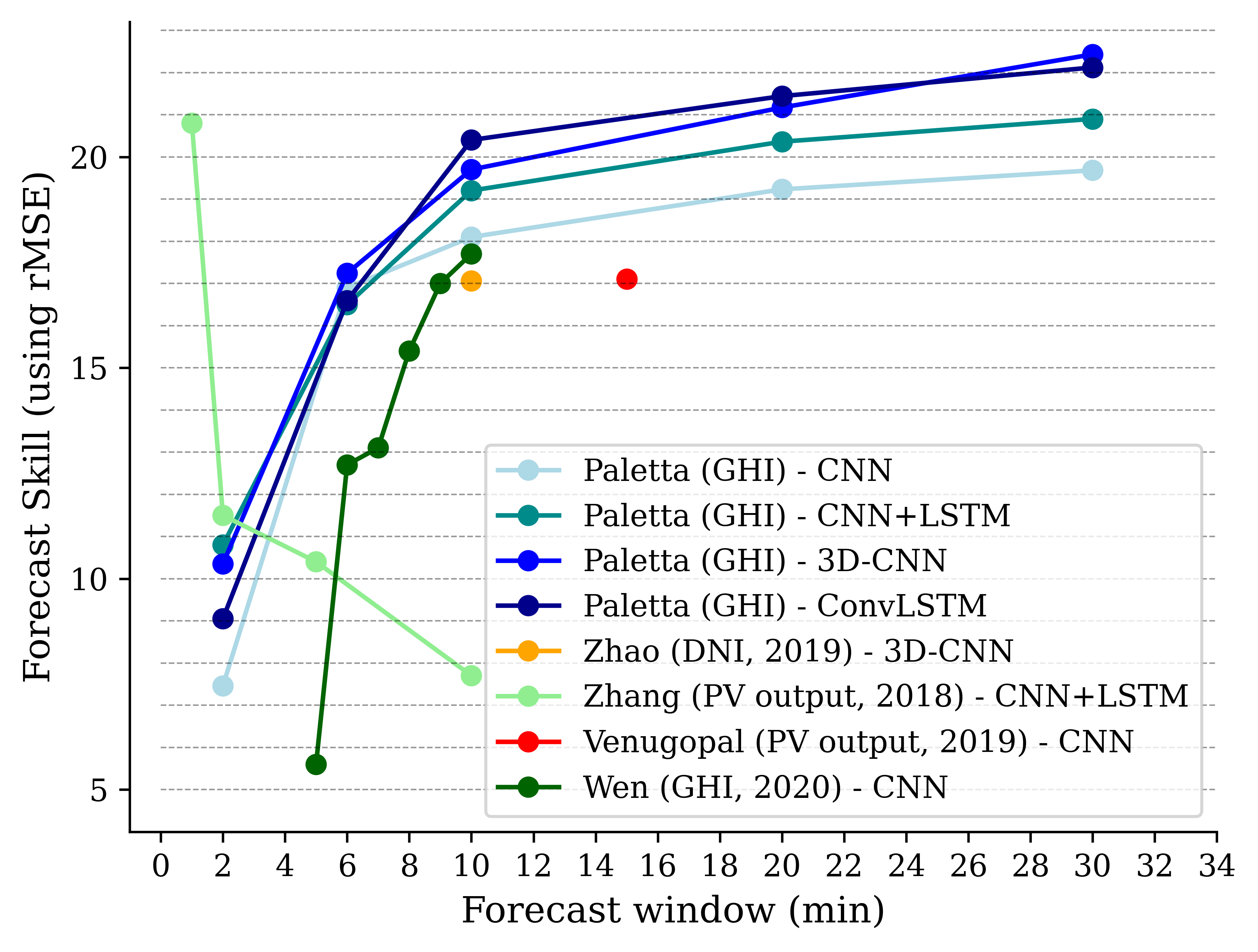}
\caption{Benchmark with other previous studies. In brackets the nature of the prediction: global horizontal irradiance (GHI), direct normal irradiance (DNI), PV output.}
\label{fig:fs_rmse_benchmark_all}
\end{figure}

Comparisons of the performance of different models tested on different datasets requires careful consideration. Not only the testing methods can be different, but also the quality, type and amount of training data (see Figure~\ref{fig:fs_training_size}) and the target of the model (i.e. GHI, DNI or PV output) may differ from one study to another~\citep{nouriEvaluationAllSky2020}. For details on the scale of differences between existing approaches, we refer the reader to the source literature. However, specific metrics such as forecast skills have been shown to better generalise over the forecast methods using different datasets~\citep{Yang2020b}. For this reason, we used the forecast skill based on the SPM using the RMSE metrics to benchmark our models with other DL approaches for varying forecast windows. The resulting comparison is presented in Figure~\ref{fig:fs_rmse_benchmark_all}.

\vspace{1\baselineskip}
Our models perform better than those of~\cite{Zhao2019}, ~\cite{Venugopal2019} and~\cite{Wen2020a} for the 5 to 15 min forecast windows. Regarding the 1 to 2 min ahead forecasts, i.e. very short term forecasts, the CNN+LSTM model developed by~\cite{Zhang2018} performs better than all four proposed models of this study. The main reasons for the difference might be that the study of~\cite{Zhang2018} was focusing on the 2 min forecast window compared to 10 min in the present study, and the temporal resolution of the corresponding dataset used to train the models was significantly higher: 15 sec compared to 2 min. Moreover, ~\cite{Zhang2018} worked directly on PV output and not on irradiance, which entails different challenges. Nevertheless, the work conducted by~\cite{Zhang2018} is promising by showing that, with adequate data, FS of DL models can exceed 20\% on very short-term forecasting.

%\bibliographystyle{elsarticle-harv} 
%\bibliography{library.bib}


%%\bibliographystyle{plainnat}
%%\bibliography{library.bib}

%\begin{thebibliography}{00}



%\end{thebibliography}
\end{document}